\documentclass{article}
\usepackage{helvet}
\usepackage{courier}
\usepackage{latexsym}

\usepackage[dvips]{graphicx}
\usepackage{subfigure}
\usepackage{psfrag}

\usepackage{epsfig}
\usepackage[ruled %vlinednoend
]{algorithm2e}

\begin{document}

\newcommand{\chr}[2][Christian]
{$^{\fboxsep=1pt\fbox{\tiny #1}}$%
\marginpar{\fbox{\parbox[t]{\marginparwidth}{\tiny #2}}}}

\newtheorem{theorem}{Theorem}
\newtheorem{obs}{Observation}
\newtheorem{corollary}{Corollary}
\newtheorem{lemma}{Lemma}
\newtheorem{definition}{Definition}
\newcommand{\proof}{\noindent {\em Proof.\ \ }}
\newcommand{\qed}{\hfill \mbox{$\Box$}\medskip}
\newcommand{\mymod}{\, \mbox{mod} \, }
\newcommand{\mydiv}{\, \mbox{div} \, }
\def\mymax{\mbox{$max$}}
\def\mymin{\mbox{$min$}}

\newcommand{\deriv}[2]{\Delta #1_#2}

\newcommand{\set}{\mathcal}
\newcommand{\myset}[1]{\ensuremath{\mathcal #1}}

\renewcommand{\theenumii}{\alph{enumii}}
\renewcommand{\theenumiii}{\roman{enumiii}}
\newcommand{\figref}[1]{Figure \ref{#1}}
\newcommand{\tref}[1]{Table \ref{#1}}
\newcommand{\And}{\wedge}
\newcommand{\myldots}{..}

\newtheorem{mydefinition}{Definition}
\newtheorem{mytheorem}{Theorem}
\newtheorem{mytheorem1}{Theorem}
\newcommand{\myproof}{\noindent {\bf Proof:\ \ }}
\newcommand{\myqed}{\hspace*{0em}\hfill\mbox{$\Box$}}

\newcommand{\atmostone}{\mbox{\sc AtMost1}}
\newcommand{\sprod}{\mbox{\sc ScalarProduct}}
\newcommand{\card}{\mbox{\sc Card}}
\newcommand{\cardpath}{\mbox{\scCardpath}}

\newcommand{\occurs}{\mbox{\sc Occurs}}
\newcommand{\propagrange}{\textit{Propag-Range}}
\newcommand{\propagroots}{\textit{Propag-Roots}}
\newcommand{\bcpropagroots}{\textit{Bound-Propag-Roots}}
\newcommand{\range}{\mbox{\sc Range}}
\newcommand{\roots}{\mbox{\sc Roots}}
\newcommand{\myiff}{\mbox{\rm iff}}
\newcommand{\alldifferent}{\mbox{\sc AllDifferent}}
\newcommand{\permutation}{\mbox{\sc Permutation}}
\newcommand{\disjoint}{\mbox{\sc Disjoint}}
\newcommand{\common}{\mbox{\sc Common}}
\newcommand{\uses}{\mbox{\sc Uses}}
\newcommand{\usedby}{\mbox{\sc UsedBy}}
\newcommand{\nvalue}{\mbox{\sc NValue}}
\newcommand{\atmostnvalue}{\mbox{\sc AtMostNValue}}
\newcommand{\among}{\mbox{\sc Among}}
\newcommand{\atmost}{\mbox{\sc AtMost}}
\newcommand{\atleast}{\mbox{\sc AtLeast}}
\newcommand{\element}{\mbox{\sc Element}}
\newcommand{\gcc}{\mbox{\sc Gcc}}
\newcommand{\opengcc}{\mbox{\sc OpenGcc}}
\newcommand{\contiguity}{\mbox{\sc Contiguity}}
\newcommand{\assignnvalues}{\mbox{\sc Assign\&NValues}}
\newcommand{\linksettobooleans}{\mbox{\sc LinkSet2Booleans}}
\newcommand{\domain}{\mbox{\sc Domain}}
\newcommand{\symalldiff}{\mbox{\sc SymAllDiff}}
\newcommand{\alldiff}{\mbox{\sc AllDifferent}}
\newcommand{\openalldiff}{\mbox{\sc OpenAllDifferent}}
\newcommand{\sumc}{\mbox{\sc Sum}}

\newcommand{\todo}[1]{{\tt (... #1 ...)}}

\SetKw{myproc}{procedure}
\SetKw{myfunc}{function}

\title{\range\ and \roots: \\Two Common Patterns \\for Specifying and
  Propagating \\Counting and Occurrence Constraints\thanks{This paper
    is a compilation and an extension of \cite{comicsIJCAI05},  \cite{bessiere06:ran},
    and \cite{bhhkwcp2006}. 
The first author was supported  by the ANR project ANR-06-BLAN-0383-02. }}

\author{
Christian Bessiere\\
LIRMM, CNRS and U. Montpellier \\
Montpellier, France \\
bessiere@lirmm.fr \and
Emmanuel Hebrard\\
4C and UCC\\
Cork, Ireland\\
e.hebrard@4c.ucc.ie \and
Brahim Hnich\\
Izmir University of Economics\\
Izmir, Turkey\\
brahim.hnich@ieu.edu.tr \and
Zeynep Kiziltan\\
Department of Computer Science\\
Univ. di Bologna, Italy \\
zeynep@cs.unibo.it \and
Toby Walsh\\
NICTA and UNSW\\
Sydney, Australia\\
tw@cse.unsw.edu.au }

\date{Keywords: Constraint programming, constraint satisfaction,
  global constraints, open global constraints, decompositions}

\maketitle
\begin{abstract}
We propose \range\ and \roots\ which are two %a simple declarative language two
common patterns
useful for specifying a wide range of counting and occurrence
constraints. We design specialised propagation algorithms for
these two patterns.
%Hence,
%Thus, for
%the 
Counting and occurrence
constraints specified using these patterns
thus directly inherit a
%we have
%polynomial
propagation algorithm. To illustrate the capabilities of the
\range\ and \roots\ constraints, we specify a number of global
constraints taken from the literature.
Preliminary
experiments demonstrate that propagating
counting and occurrence constraints using these two patterns leads
to a small loss in performance when compared to specialised
global constraints and is competitive with alternative decompositions
using elementary constraints.
\end{abstract}

\section{Introduction}

Global constraints are central to the success of constraint
programming \cite{handbook}. Global constraints allow users to specify patterns
that occur in many problems, and to exploit efficient and
effective propagation algorithms for pruning the search space. Two
common types of global constraints are counting and occurrence
constraints.
Occurrence constraints place restrictions on the occurrences of
particular values. For instance, we may wish to ensure that no
value used by one set of variables occurs in a second set.
Counting constraints, on the other hand, restrict the number of
values or variables meeting some condition. For example, we may
want to limit the number of distinct values assigned to a set of
variables. Many different counting and occurrences constraints
have been proposed to help model a wide range of problems,
especially those involving resources (see, for example,
\cite{regin1,beldiceanu5,regin2,bcp01,beldiceanu7}).

In this paper, we will show that many such constraints can be
specified  by means of two new global constraints, \range\ and
\roots\, together with some standard elementary constraints like
subset and set cardinality. 
%***CB: added on Jan 26, 2009
These two new global constraints capture the
familiar notions of \emph{image} and \emph{domain} of a
function. Understanding such notions does not require a strong
background in constraint programming. A basic mathematical background
is sufficient to understand these constraints and use them to
specify other global constraints. 
We will show, for example, that
\range\ and \roots\ are versatile enough to allow specification of
\emph{open}  global constraints, a recent kind of global constraints
for  which the set of variables involved is not known in advance. 

%This s
Specifications made  with \range\ and \roots\ constraints 
%***end of add
%is %also
are executable. We show that efficient propagators can be
designed for the \range\ and \roots\ constraints. We  give an
efficient algorithm for propagating the \range\ constraint based
on a flow algorithm.
We also prove that it is intractable to propagate the
\roots\ constraint completely. We therefore propose a
decomposition of the \roots\ constraint that can propagate it
partially in linear time. This decomposition does not destroy the
global nature of the \roots\ constraint as in many situations met
in practice, it prunes all possible values. The proposed
propagators can easily be incorporated into a constraint toolkit.

We show that specifying a global constraint using \range\ and
\roots\    provides us with an {\em reasonable} method to
propagate counting and occurrence constraints. There are three
possible situations. In the first,
the global nature of the \range\ and \roots\ constraints is enough
to capture the global nature of the given counting or occurrence
constraint, and propagation is not hindered. In the second
situation, completely propagating the counting or occurrence
constraint is NP-hard. We must accept some loss of propagation if
propagation is to be tractable. Using \range\ and \roots\ is then
one means to propagate the counting or occurrence constraint
partially. In the third situation, the global constraint can be
propagated completely in polynomial time but using \roots\ and
\range\ hinders propagation.
In this case,
if we want to achieve
full propagation, we need to develop a specialised propagation
algorithm.

We also show that decomposing occurrence constraints and counting
constraints using the \range\ and \roots\ constraints performs
well in practice. Our experiments on random %binary
CSPs and a on real world problem from
CSPLib  demonstrate that propagating counting and occurrence
constraints using the \range\ and \roots\ constraints leads to
a small loss in performance when compared to specialised global
constraints and is competitive with alternative decompositions
into more elementary constraints.

The rest of the paper is organised as follows. Section
\ref{sec:preli} gives the formal background. Section
\ref{sec:lang} defines the \range\ and \roots\ constraints and
gives a couple of  examples to illustrate how global constraints
can be decomposed using these two constraints. In Section
\ref{sec:range}, we propose a polynomial algorithm for the \range\
constraint. 
In Section
\ref{sec:roots}, we give a complete theoretical analysis of the
\roots\ constraint and our decomposition of it, and
we discuss implementation details. Section \ref{sec:catalog} gives
many examples of counting and occurrence constraints that can be
specified using the \range\ and \roots\ constraints. Experimental
results are presented in Section \ref{sec:exp}. Finally, we end with
conclusions in Section \ref{sec:concl}.

\section{Formal background}
\label{sec:preli}

A constraint satisfaction problem consists of a set of variables,
each with a finite domain of values, and a set of constraints
specifying allowed combinations of values for subsets of
variables.
We use capitals for variables (e.g. $X$, $Y$ and $S$), and lower
case for values (e.g. $v$ and $w$).
We write ${D}(X)$ for the domain of a variable $X$.
A solution is an assignment of values to the variables satisfying
the constraints. A variable is \emph{ground} when it is assigned a
value. We consider both \emph{integer} and \emph{set} variables.
A set variable $S$ is often represented by
its lower bound $lb(S)$ which contains the definite elements (that must belong to the set) % of the set
and an upper bound $ub(S)$ which also contains the %definite and
potential elements (that may or may not belong to the set).

Constraint solvers typically explore partial assignments enforcing
a local consistency property using either specialised or general
purpose propagation algorithms.
Given a constraint $C$, 
a \emph{bound support} on $C$ is a tuple %$\tau$ %on $var(C)$
that assigns to each integer variable a value between its minimum
and  maximum, and to each set variable a set between its lower and
upper bounds which satisfies $C$. A bound support in which each
integer variable is assigned a value in its domain is called a
\emph{hybrid support}.
If $C$ involves only integer variables, a hybrid
support is %called
a \emph{support}.  %on $C$ is a tuple %$\tau$ %on $var(C)$
A value %$v$
(resp. set of values) %$s$)
for an integer variable %$X_i$
(resp. set variable) %$S_j$)
is \emph{bound} or \emph{hybrid consistent with} $C$ iff there
exists
a bound or  hybrid  support %$\tau$
assigning this value (resp. set of values) to this variable.
A constraint $C$ is \emph{bound consistent} (\emph{BC}) iff for
each integer
variable $X_i$, %in $var(C)$,
its minimum and maximum values belong to a bound support,
and for each set variable $S_j$, %in $var(C)$,
the values in $ub(S_j)$  belong to $S_j$ in  at least one bound
support %on $C$,
and the values in $lb(S_j)$ are those from $ub(S_j)$ that
belong to $S_j$ in all bound supports. % on $C$.
A constraint $C$ is \emph{hybrid consistent} (\emph{HC}) iff for
each integer
variable $X_i$, %in $var(C)$,
every value in $D(X_i)$ belongs to a hybrid support,
and for each set variable $S_j$, %in $var(C)$,
the values in $ub(S_j)$  belong to $S_j$ in  at least one hybrid
support, %on $C$
and the values in $lb(S_j)$ are those from $ub(S_j)$ that
belong to $S_j$ in all hybrid supports. % on $C$.
A constraint $C$ involving only integer variables is
\emph{generalised
  arc  consistent} (\emph{GAC})
iff for each
variable $X_i$, %in $var(C)$,
every value in $D(X_i)$ belongs to a support.
If all variables in $C$ are integer variables, hybrid consistency
reduces to generalised arc consistency, and if all variables in
$C$ are set variables, hybrid consistency reduces to bound
consistency.

To illustrate these different concepts, consider the constraint
$C(X_1,X_2,T)$ that holds iff the set variable $T$ is assigned
exactly the values used by the integer variables $X_1$ and $X_2$.
Let $D(X_1)=\{1,3\}$, $D(X_2)=\{2,4\}$, $lb(T)=\{2\}$ and
$ub(T)=\{1,2,3,4\}$. BC does not remove any value since all
domains are already bound consistent (value 2 was considered as
possible for $X_1$ because BC deals with bounds). On the other
hand, HC removes 4 from $D(X_2)$ and from $ub(T)$ as there does
not exist any tuple satisfying $C$ in which $X_2$ does not take
value 2.%Finally,

\newcommand{\tighter}{\mbox{$\preceq$}}
\newcommand{\stighter}{\mbox{$\prec$}}
\newcommand{\incomparable}{\mbox{$\bowtie$}}
\newcommand{\equivalent}{\mbox{$\equiv$}}

We will compare local consistency properties applied to (sets of)
logically equivalent constraints, $c_1$ and $c_2$.
As in \cite{debruyne1}, %we say that
a local consistency property $\Phi$ on $c_1$ is as strong as
$\Psi$ on $c_2$ %(written $\Phi(c_1) \tighter \Psi(c_2)$)
iff, given any domains, if $\Phi$ holds on $c_1$ then $\Psi$ holds
on $c_2$;
$\Phi$ on $c_1$ is stronger
than $\Psi$ on $c_2$ %(written $\Phi(c_1) \stighter \Psi(c_2)$)
iff %$\Phi(c_1) \tighter \Psi(c_2)$ but not $\Psi(c_2) \tighter \Phi(c_1)$
$\Phi$ on $c_1$ is as strong as $\Psi$ on $c_2$ but not vice
versa;
$\Phi$ on $c_1$ is equivalent to
$\Psi$ on $c_2$ %(written $\Phi(c_1) \equivalent \Psi(c_2)$)
iff
$\Phi$ on $c_1$ is as strong as $\Psi$ on $c_2$ and vice versa;
$\Phi$ on $c_1$ is incomparable to
$\Psi$ on $c_2$ %(written $\Phi(c_1) \equivalent \Psi(c_2)$)
iff
$\Phi$ on $c_1$ is not as strong as $\Psi$ on $c_2$ and vice
versa.

A total function $\set{F}$ from  a source set $\set{S}$ into a target set
$\set{T}$ is denoted by $\set{F}: \set{S} \longrightarrow
\set{T}$. The set of all elements in $\set{S}$ that have the same
image
$j \in \set{T}$ is %denoted by
$\set{F}^{-1}(j)=\{i: %i \in {S},
\set{F}(i)=j\}$. The image of a set ${S} \subseteq \set{S}$
under $\set{F}$ is %denoted by
$\set{F}({S})=\bigcup_{i \in {S}} \set{F}(i)$, whilst the domain of
a set ${T}
\subseteq \set{T}$ under $\set{F}$ is %denoted by
$\set{F}^{-1}({T})=\bigcup_{j \in {T}} \set{F}^{-1}(j)$.
Throughout, we will view a set of integer variables, $X_1$ to $X_n$
as a function $\set{X}: \{1,\myldots,n\} \longrightarrow
\bigcup_{i=1}^{i=n}
\set{D}(X_i)$. % where $D(X_i)$ denotes the domain of $X_i$.
That is, %we see
$\set{X}(i)$ is the value of $X_i$.

\section{Two useful patterns: \range\ and \roots}
\label{sec:lang}

Many counting and occurrence constraints can be specified using
simple non-global constraints over integer variables (like $X \leq
m$), simple non-global constraints over set variables (like $S_1
\subseteq S_2$ or $|S|=k$) available in most
constraint solvers, and \emph{two special global constraints}
acting on sequences of variables: \range\ and \roots. 
%***CB: added on Jan, 26, 2009
 \range\  captures the notion of \emph{image}  of a
function and \roots\ captures the notion of \emph{domain}. 
%***end of add
%This
Specification with \range\ and \roots\ is executable. 
It permits us to decompose  %each
other
global constraints into more primitive constraints.

Given a function $\set{X}$ representing a set of integer
variables, $X_1$ to $X_n$, the \range\ constraint holds iff a set
variable $T$ is the image of another set variable $S$ under
$\set{X}$.
$$\range([X_1,\myldots,X_n],S,T) \ \ \myiff\ \ T = \set{X}({S})
\ \ (\textrm{that is}, T= \{X_i~|~i \in S\} )$$
The \roots\ constraint holds iff a set variable $S$ is the domain of
the another set variable $T$ under $\set{X}$.

$$\roots([X_1,\ldots,X_n],S,T) \ \myiff\ \ {S}=\set{X}^{-1}({T})
\ \ (\textrm{that is}, S= \{i~|~X_i \in T\} )$$
\range\ and \roots\ are not exact inverses. A \range\ constraint can
hold, but the corresponding \roots\ constraint may not, and vice
versa.
For instance, $\range([1,1],\{1\},\{1\})$ holds but not
$\roots([1,1],\{1\},\{1\})$ since $\set{X}^{-1}(1)=\{1,2\}$, and
$\roots([1,1,1],\{1,2,3\},\{1,2\})$ holds but not
$\range([1,1,1],\{1,2,3\},\{1,2\})$ as no $X_i$ is assigned to
$2$.

Before showing how to propagate \range\ and \roots\ efficiently,
we give two examples that illustrate how some counting and
occurrence global constraints from \cite{beldiceanu3} can be
specified using \range\ and \roots .

The \nvalue\ constraint  counts the number of distinct values
used by a sequence of variables 
\cite{pachet1,comicsNVALUE,comicsCONSTRAINTS06}.
$\nvalue([X_1,\myldots,X_n],N)$ holds iff $N=|\{ X_i \ | \ 1 \leq
i \leq n\}|$. A  way to implement this constraint is
with a \range\ constraint:

%\vspace{-0.7em}
\begin{eqnarray*}
& & \nvalue([X_1,\myldots,X_n],N)  \ \ \myiff    \\
& & \range([X_1,\myldots,X_n],\{1,\myldots,n\},T) \ \And \ |T|=N
\end{eqnarray*}
%\vspace{-1em}

The \atmost\  constraint is one of the oldest global constraints
\cite{vanhdevICLP91}. The
\atmost\ constraint puts an upper bound on the number of variables
using a particular value.
$\atmost([X_1,\myldots,X_n],d,N)$ holds
iff $|\{ i \ | \ X_i=d\}| \leq N$.
It  can be decomposed using a \roots\
constraint.

\begin{eqnarray*}
& & \atmost([X_1,\myldots,X_n],d,N)  \ \ \myiff    \\
& & \roots([X_1,\myldots,X_n],S,\{d\}) \ \And \ |S| \leq N
\end{eqnarray*}

These two examples show that it can be quite simple to decompose
global constraints using \range\ and \roots. As we will show
later, some other global constraints will require the use of both
\range\ and \roots\  in the same decomposition. The next sections
show how \range\ and \roots\ can be propagated efficiently.% so as

\section{Propagating the \range\ constraint}
\label{sec:range}

Enforcing hybrid consistency on the \range\ constraint is
polynomial. This can be done using a maximum network flow problem.
In fact, the \range\ constraint can be decomposed using a global
cardinality constraint (\gcc) for which propagators based on flow
problems already exist \cite{regin2,quietalCP04}.
But the \range\ constraint does
not need the whole power of maximum network flow problems, and thus
HC can be enforced on it at a lower cost than that of calling a
\gcc\ propagator. In this section, we propose an efficient way to
enforce HC on \range. %\ using a maximum network flow problem.
To simplify the presentation, the use of the flow is limited to a
constraint that performs only part of the work needed for
enforcing HC on \range. This constraint, that we name
$\occurs([X_1,\ldots,X_n],T)$, ensures that  all the values in the
set variable $T$ are used by the integer variables $X_1$ to $X_n$:
$$
\occurs([X_1,\ldots,X_n], T) \ \ \myiff \  \ T \subseteq
\set{X}({\{1..n\}})
\ \ (\textrm{that is}, T\subseteq \{X_i~|~i \in 1..n\} )
$$

We first present an  algorithm for achieving HC on \occurs\
(Section \ref{sec:occurs}), and then use this to propagate the
\range\ constraint (Section \ref{sec:hcrange}).

\subsection{Hybrid consistency on \occurs}
\label{sec:occurs}

We achieve HC on $\occurs([X_1,\ldots,X_n],T)$ using a network flow.

\subsubsection{Building the network flow}

\begin{figure}[tbp]
 \centering
\psfrag{1}{$1$} \psfrag{2}{$2$} \psfrag{3}{$3$} \psfrag{4}{$4$}
\psfrag{s}{$s$} \psfrag{t}{$t$} \psfrag{z1}{$z_1$}
\psfrag{z2}{$z_2$} \psfrag{z3}{$z_3$} \psfrag{z4}{$z_4$}
\psfrag{x1}{$x_1$} \psfrag{x2}{$x_2$} \psfrag{x3}{$x_3$}
 \includegraphics[width=5cm]{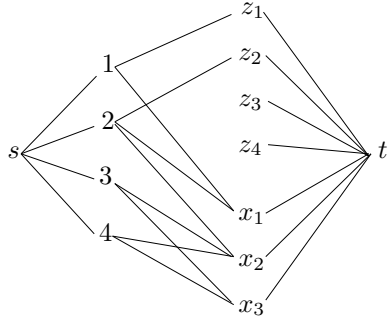}
 \caption{Unit capacity network of the constraint
   $C=\occurs([X_1,X_2,X_3],T)$  with  $D(X_1)=\{1,2\}$,
   $D(X_2)=\{2,3,4\}$, $D(X_3)=\{3,4\}$,  $lb(T)=\{3,4\}$ and
   $ub(T)=\{1,2,3,4\}$. Arcs are directed from left to right. \label{fig:flow1}}
\end{figure}

We use a unit capacity network \cite{ahujaBOOK93} in which
capacities between two nodes can only be 0 or 1. This is represented
by a directed graph where an arc from node $x$ to node $y$ means
that a maximum flow of 1 is allowed between $x$ and $y$ while the
absence of an arc means that the maximum flow allowed is~0. The unit
capacity network $G_{C}=(N,E)$ of the constraint
$C=\occurs([X_1,\ldots,X_n],T)$ is built in the following way.
$N=\{s\}\cup N_1\cup N_2\cup\{t\}$, where $s$ is a source node, $t$
is a sink node, $N_1=\{v~|~ v \in \bigcup D(X_i)\}$ and
$N_2=\{z_v~|~v \in \bigcup D(X_i)\}\cup\{x_{i}~|~i\in [1..n]\}$. The
set of arcs $E$ is as follows:
$$
E= (\{s\}\times N_1) \cup\{(v,z_v), \forall v\notin
lb(T)\}\cup\{(v,x_{i})~|~v\in D(X_i)\}\cup (N_2\times\{t\})
$$
$G_C$ is quadripartite, i.e., $E\subseteq(\{s\}\times N_1)\cup
(N_1\times N_2)\cup(N_2\times\{t\})$. In Fig. \ref{fig:flow1}, we
depict the network $G_C$ of the constraint
$C=\occurs([X_1,X_2,X_3],T)$  with $D(X_1)=\{1,2\}$,
$D(X_2)=\{2,3,4\}$,  $D(X_3)=\{3,4\}$,  $lb(T)=\{3,4\}$ and
$ub(T)=\{1,2,3,4\}$. The intuition behind this graph is that when
a flow uses an arc from a node $v$ to a node $x_i$ this means that
$X_i$ is assigned $v$, and when a flow uses the arc $(v,z_v)$ this
means that $v$ is not necessarily used by the
$X_i$'s.\footnote{Note that in our presentation of the graph,
  the edges go from the nodes
  representing the values to the nodes representing the
  variables. This is the opposite to the direction used in the
  presentation of network
  flows for propagators of the \alldifferent\ or \gcc\
  constraints \cite{regin1,regin2}. }
In Fig. \ref{fig:flow1} nodes 3 and 4 are linked only to nodes $x_2$
and $x_3$, that is, values 3 and 4 must necessarily be taken by one
of the variables $X_i$ (3 and 4 belong to $lb(T)$). On the contrary,
nodes 1 and 2 are also linked to nodes $z_1$ and $z_2$ because
values 1 and 2 do not have to be taken by a $X_i$ (they are not in
$lb(T)$).

\begin{figure}[tbp]
 \centering
\psfrag{1}{$1$} \psfrag{2}{$2$} \psfrag{3}{$3$} \psfrag{4}{$4$}
\psfrag{s}{$s$} \psfrag{t}{$t$} \psfrag{z1}{$z_1$}
\psfrag{z2}{$z_2$} \psfrag{z3}{$z_3$} \psfrag{z4}{$z_4$}
\psfrag{x1}{$x_1$} \psfrag{x2}{$x_2$} \psfrag{x3}{$x_3$}
 \includegraphics[width=5cm]{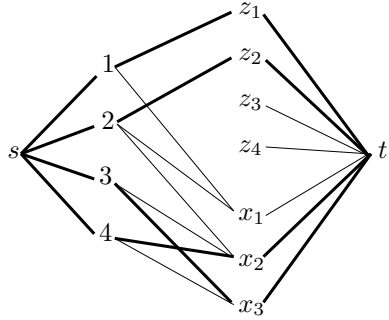}
 \caption{A maximum flow for the network of
   Fig. \ref{fig:flow1}. Bold arcs are those that belong to the
   flow. Arcs are directed from left to right.\label{fig:flow2}}
\end{figure}

In the particular case of unit capacity networks, a flow is any
set $E'\subseteq E$: any arc in $E'$ is assigned 1 and the arcs in
$E\setminus E'$ are assigned~0.
A \emph{feasible} flow from $s$ to $t$ in $G_C$ is a subset $E_f$
of $E$ such that $\forall n\in N\setminus \{s,t\}$ the number of
arcs of $E_f$ entering $n$ is equal to the number of arcs of $E_f$
going out of $n$, that is, $|\{(n',n)\in E_f\}|=|\{(n,n'')\in
E_f\}|$. The value of the flow $E_f$ from $s$ to $t$, denoted
$val(E_f,s,t)$,
is %the amount of commodity going from $s$ to $t$, i.e.,
$val(E_f,s,t)=|\{n~|~(s,n)\in E_f\}|$. A \emph{maximum} flow from
$s$ to $t$ in $G_C$ is a feasible flow $E_M$ such that there does
not exist a feasible flow $E_f$, with $val(E_f,s,t)>val(E_M,s,t)$.
A maximum flow for the network of Fig. \ref{fig:flow1} is given in
Fig. \ref{fig:flow2}.
By construction a feasible flow cannot have a value greater than
$|N_1|$ and cannot contain two arcs entering a node $x_i$ from
$N_2$. Hence, we can  define a function $\varphi$ linking feasible
flows and partial instantiations on the $X_i$'s. Given any feasible
flow $E_f$ from $s$ to $t$ in $G_C$,
$\varphi(E_f)=\{(X_i,v)~|~(v,x_i)\in E_f\}$. The maximum flow in
Fig. \ref{fig:flow2} corresponds to the instantiation $X_2=4,
X_3=3$. The way $G_C$ is built induces the following theorem.

\begin{mytheorem}\label{theo:occur:hc}
Let $G_C=(N,E)$ be the capacity network of a constraint
$C=\occurs([X_1,\ldots,X_n],T)$.
\begin{enumerate}
\item A value  $v$ in the domain $D(X_i)$ for some $i\in [1..n]$ is HC
  iff there exists a % (maximum)
  flow $E_f$ from $s$ to $t$ in $G_C$ with
  $val(E_f,s,t)=|N_1|$ and $(v,x_i)\in E_f$
\item If the $X_i$'s are HC, $T$ is HC
  iff $ub(T)\subseteq \bigcup_i D(X_i)$
\end{enumerate}
\end{mytheorem}
\proof (1.$\Rightarrow$) Let $I$ be a solution  for $C$ with
$(X_i,v)\in I$. Build the following flow $H$: Put $(v,x_i)$ in
$H$; $\forall w\in I[T],w\neq v$, take a variable $X_j$ such that
$(X_j,w)\in I$ (we know there is at least one since $I$ is
solution), and put $(w,x_j)$ in $H$; $\forall w'\notin I[T],
w'\neq v$, add $({w'},z_{w'})$ to $H$. Add to $H$ the edges from
$s$ to $N_1$ and from $N_2$ to $t$ so that we obtain a feasible
flow. By construction, all $w\in N_1$ belong to an edge of $H$.
So, $val(H,s,t)=|N_1|$ and $H$ is a maximum flow with $(v,x_i)\in
H$.

(1.$\Leftarrow$) Let $E_M$ be a  flow from $s$ to $t$ in $G_C$
with $(v,x_i)\in E_M$ and $val(E_M,s,t)$ $=|N_1|$. By construction
of $G_C$, we are guaranteed that all nodes in $N_1$ belong to an
arc in $E_M\cap (N_1\times N_2)$, and that for every value $w\in
lb(T)$, $\{y~|~(w,y)\in E\}\subseteq\{x_i~|~i\in [1..n]\}$. Thus,
for each $w\in lb(T), \exists X_j~|~(X_j,w)\in\varphi(E_M)$.
Hence, any  extension of $\varphi(E_M)$ where each unassigned
$X_j$ takes any value in $D(X_j)$ and $T=lb(T)$ is a solution of
$C$ with $X_i=v$.

(2.$\Rightarrow$) If $T$ is HC, all values in $ub(T)$ appear in at
least one solution tuple. Since $C$ ensures that $T\subseteq
\bigcup_i\{X_i\}$, $ub(T)$ cannot contain a value appearing in
none of the $D(X_i)$.

(2.$\Leftarrow$) Let $ub(T)\subseteq \bigcup_i D(X_i)$. Since all
$X_i$'s are HC, we know that each value $v$ in $\bigcup_i D(X_i)$
is taken by some $X_i$ in at least one solution tuple $I$. Build
the tuple $I'$ so that $I'[X_i]=I[X_i]$ for each $i\in [1..n]$ and
$I'[T]=I[T]\cup\{v\}$. $I'$ is still solution of $C$. So, $ub(T)$
is as tight as it can be wrt HC. In addition, since all $X_i$'s
are HC, this means that in every solution tuple $I$, for each
$v\in lb(T)$ there exists $i$ such that $I[X_i]=v$. So, $lb(T)$ is
HC. \qed

\begin{figure}[tbp]
 \centering
\psfrag{1}{$1$} \psfrag{2}{$2$} \psfrag{3}{$3$} \psfrag{4}{$4$}
\psfrag{s}{$s$} \psfrag{t}{$t$} \psfrag{z1}{$z_1$}
\psfrag{z2}{$z_2$} \psfrag{z3}{$z_3$} \psfrag{z4}{$z_4$}
\psfrag{x1}{$x_1$} \psfrag{x2}{$x_2$} \psfrag{x3}{$x_3$}
 \includegraphics[width=5cm]{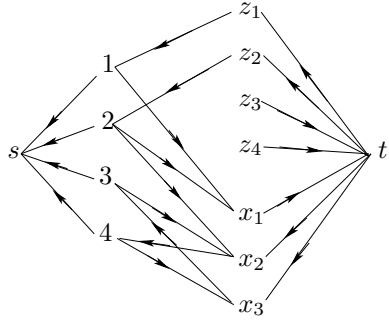}
 \caption{Residual graph obtained from the network in
   Fig. \ref{fig:flow1} and the maximum flow in Fig. \ref{fig:flow2}.
   \label{fig:flow3}}
\end{figure}

Following Theorem \ref{theo:occur:hc}, we need a way to check
which edges belong to a maximum flow. \emph{Residual graphs} are
useful for this task. Given a unit capacity network $G_C$ and a
maximal flow $E_M$  from $s$ to $t$ in $G_C$, the residual graph
$R_{G_C}(E_M)=(N,E_R)$ is the directed graph obtained from $G_C$
by reversing all arcs belonging to the maximum flow $E_M$; that
is, $ E_R=\{(x,y)\in E\setminus E_M\}\cup\{(y,x)~|~(x,y)\in E\cap
E_M\} $. Given the network $G_C$ of Fig. \ref{fig:flow1} and the
maximum flow $E_M$ of Fig. \ref{fig:flow2}, $R_{G_C}(E_M)$ is
depicted in Fig. \ref{fig:flow3}.
%Thanks to a theorem from Berge \cite{berge70}, we know that
Given a maximum flow $E_M$ from $s$ to $t$ in $G_C$, given
$(x,y)\in N_1\times N_2\cap E\setminus E_M$, there exists a
maximum flow containing $(x,y)$ iff $(x,y)$ belongs to a cycle in
$R_{G_C}(E_M)$ \cite{schrijver03}. Furthermore, finding all the
arcs $(x,y)$ that do not belong to a cycle in a graph can be
performed by building the strongly connected components  of the
graph. We see in Fig. \ref{fig:flow3} that the arcs $(1,x_1)$ and
$(2,x_1)$ belong to a cycle. So, they belong to some maximum flow
and $(X_1,1)$ and $(X_1,2)$ are hybrid consistent. $(2,x_2)$ does
not belong to any cycle. So, $(X_2,2)$ is not HC.

\subsubsection{Using the network flow for achieving HC on \occurs}

We now have all the tools for achieving HC on any \occurs\
constraint.
We first build $G_C$.
We  compute a maximum flow $E_M$ from $s$ to $t$ in $G_C$; if
$val(E_M,s,t)<|N_1|$,  we fail. Otherwise
%\item.
we compute $R_{G_C}(E_M)$,
%\item
build the strongly connected components  in $R_{G_C}(E_M)$, and
%\item
remove from $D(X_i)$ any value $v$ such that $(v,x_i)$ belongs to
neither $E_M$ nor to a strongly connected component in
$R_{G_C}(E_M)$. Finally,
%\item
we set $ub(T)$ to $ub(T)\cap \bigcup_i D(X_i)$.
%\end{enumerate}
Following Theorem \ref{theo:occur:hc} and properties of residual
graphs, this algorithm  enforces HC on $\occurs([X_1,..,X_n],T)$.
\\

\noindent \emph{Complexity.}
Building $G_C$ % (step 1.)
is in $O(nd)$ where $d$ is the maximum domain size. We need then
to find a
maximum flow $E_M$ in $G_C$% (step 2.)
. This can be done in two sub-steps. First, we use the arc
$(v,z_v)$  for each $v\notin lb(T)$ (in $O(|\bigcup_i D(X_i)|)$).
Afterwards, we compute a maximum flow on the subgraph composed of
all paths traversing nodes $w$ with $w\in lb(T)$ (because there is
no arc $(w,z_w)$ in $G_C$ for such $w$). The complexity of finding
a maximum flow in a unit capacity network is in $O(\sqrt{k}\cdot
e)$ if $k$ is the number of nodes and $e$ the number of edges.
This gives a
complexity in $O(\sqrt{|lb(T)|}\cdot |lb(T)|\cdot n)$ %
%\footnote{This is a simplification of the more accurate
%  bound  $O(\sqrt{|lb(T)|}\cdot n\cdot min(d,|lb(T)|))$.}
for this second sub-step. Building the residual graph % (step 3.)
and
computing the strongly connected components % (step 4.)
is in $O(nd)$. Extracting the HC
domains for the $X_i$'s is direct% (step 5.)
. There remains to
compute BC on $T$% (step 6.)
,  which takes $O(nd)$.
Therefore, the total complexity is in $O(nd+n\cdot|lb(T)|^{3/2})$.\\

\noindent \emph{Incrementality.} In constraint solvers, constraints
are usually \emph{maintained} in a locally consistent state after
each modification (restriction) of the domains of the variables. It
is thus interesting  to consider the total complexity of
maintaining HC on \occurs\ after an arbitrary number of restrictions
on the domains (values removed from $D(X_i)$ and $ub(T)$, or added
to $lb(T)$) as we descend a branch of a backtracking
search tree. Whereas some constraints are completely incremental
(i.e., the total complexity after any number of restrictions is the
same as the complexity of one propagation), this is not the case for
constraints based on flow techniques like \alldifferent\ or \gcc\
\cite{regin1,regin2}.
%, \emph{symmetric cardinality} \cite{kockreCPAIOR04}, etc.
They potentially require the computation of a new maximum
flow after each modification. Restoring a maximum flow from one
that lost $p$ edges is in $O({p}\cdot e)$. If values are removed
one by one ($nd$ possible times),  and if each removal affects the
current maximum flow, the overall complexity over a sequence of
restrictions on $X_i$'s, $S$, $T$, is in $O(n^2d^2)$.

\subsection{Hybrid consistency on \range}
\label{sec:hcrange}

Enforcing HC on $\range([X_1,\ldots,X_n],S,T)$ can be done by
decomposing it as an \occurs\ constraint on new variables $Y_i$
and some channelling constraints (\cite{cheng99}) linking  $T$ and
the $Y_i$'s to $S$ and the $X_i$'s. Interestingly, 
we do not need to \emph{maintain} HC on the decomposition but
just need to propagate the constraints  in \emph{one pass}.

The algorithm \propagrange, enforcing HC on the \range\
constraint, is presented in Algorithm \ref{algo:range}. In line
\ref{algo:init}, a special encoding is built, where a $Y_i$ is
introduced for each $X_i$ with index in $ub(S)$. The domain of a
$Y_i$ is the same as that of $X_i$ plus a dummy value. The dummy
value works as a flag. If \occurs\ prunes it from $D(Y_i)$ this
means that $Y_i$ is necessary in \occurs\ to cover $lb(T)$. Then,
$X_i$ is also necessary to cover $lb(T)$ in \range.
In line  \ref{algo:occ}, HC on \occurs\ removes a value from a
$Y_i$ each time it contains other values that are necessary to
cover $lb(T)$ in every solution tuple. HC also removes values from
$ub(T)$ that cannot be covered by any $Y_i$ in a solution.
%Line \ref{algo:inc} propagates deletions from $T'$ to $T$.
Line \ref{algo:Sbounds} updates the bounds of $S$ and the domain
of  $Y_i$'s. Finally, in line \ref{algo:channelling}, the
channelling constraints between $Y_i$ and $X_i$ propagate removals
on $X_i$ for each $i$ which belongs to $S$ in all solutions.

\begin{algorithm}[tbp]
\begin{footnotesize}
\hspace*{-\algomargin}
\myproc{Propag-Range}{$([X_1,\ldots,X_n],S,T)$}\; \lnl{algo:init}
Introduce the set of integer variables $Y=\{Y_i~|~i \in ub(S)\}$,
\\with $D(Y_i)= D(X_i)\cup \{dummy\}$\;
%\lnl{algo:init2}
%Introduce a set variable $T'$ with $lb(T')=lb(T)$ and
%$ub(T')=ub(T)\cup\{dummy\}$\;
\lnl{algo:occ} Achieve hybrid consistency on the constraint
$\occurs(Y,T)$\;
%\lnl{algo:inc}
%Achieve BC on the constraint $T\subseteq T'$\;
\lnl{algo:Sbounds} Achieve hybrid consistency on the constraints
  $i \in S \leftrightarrow Y_i \in T$,  for all $Y_i \in Y$\;
\lnl{algo:channelling}
Achieve GAC on the  %channelling
constraints $(Y_i=dummy)\lor (Y_i = X_i)$,
% $X_i=v\to Y_i=v$,
for all $Y_i \in  Y$\;
\end{footnotesize}
\caption{\FuncSty{Hybrid consistency on
$\range$\label{algo:range}} }
\end{algorithm}

\begin{mytheorem}\label{theo:fcd}
The algorithm \propagrange\ is a correct %sound and complete
algorithm for enforcing %hybrid consistency
HC on \range, that runs in $O(nd+n\cdot|lb(T)|^{3/2})$ time, where
$d$ is the maximal size of $X_i$ domains.
\end{mytheorem}

\proof \emph{Soundness}. A value $v$ is removed from $D(X_i)$ in
line \ref{algo:channelling} if it is removed from $Y_i$ together
with $dummy$ in lines  \ref{algo:occ} or \ref{algo:Sbounds}. If a
value $v$ is removed from $Y_i$ in line \ref{algo:occ}, this means
that any tuple on variables in $Y$ covering $lb(T)$ requires that
$Y_i$ takes a value from $D(Y_i)$ other than $v$. So, we cannot
find a solution of \range\ in which $X_i=v$ since $lb(T)$ must be
covered as well. A value $v$ is removed from $D(Y_i)$ in line
\ref{algo:Sbounds} if $i\in lb(S)$ and $v\not\in ub(T)$. In this
case, \range\ cannot be satisfied by a tuple where $X_i=v$.
%Hence, after \propagrange, the $X_i$ are GAC.
If a value $v$ is removed from $ub(T)$ in line
% \ref{algo:inc}, this is because $v$ has been removed from $ub(T')$
% in line
\ref{algo:occ},
% (by construction of $T'$). Thus,
none of the tuples of values for variables in $Y$ covering $lb(T)$
can cover $v$ as well. Since variables in $Y$ duplicate variables
$X_i$ with index in $ub(S)$, there is no hope to satisfy \range\
if $v$ is in $T$. Note that $ub(T)$ cannot be modified in line
\ref{algo:Sbounds} since $Y$ contains only variables $Y_i$ for
which $i$ was in $ub(S)$. If a value $v$ is added to $lb(T)$ in
line \ref{algo:Sbounds},
% ($lb(T)$ cannot be modified in line \ref{algo:inc})
this is because there exists $i$ in $lb(S)$ such that $D(Y_i)\cap
ub(T)=\{v\}$. Hence, $v$ is necessarily in $T$ in all solutions of
\range. An index $i$ can be removed from $ub(S)$ only in line
\ref{algo:Sbounds}. This happens when the domain of $Y_i$ does not
intersect $ub(T)$.  In such a case, this is evident that a tuple
where $i\in S$ could not satisfy \range\ since $X_i$ could not
take a value in $T$. Finally, if an index $i$ is added to $lb(S)$
in line \ref{algo:Sbounds}, this is because $D(Y_i)$ is included
in $lb(T)$, which means that the dummy value has been removed from
$D(Y_i)$ in line \ref{algo:occ}. This means that $Y_i$  takes a
value from $lb(T)$  in all solutions of \occurs. $X_i$ also has to
take a value from $lb(T)$ in all solutions of \range.

%\medskip
\noindent\emph{Completeness} %(Sketch).
Suppose that a value $v$ is
not pruned from  $D(X_i)$ after line \ref{algo:channelling} of
\propagrange. If $Y_i\in Y$, we know that after line
\ref{algo:occ} there was an instantiation $I$ on $Y$ and $T$,
solution of \occurs\ with $I[Y_i]=v$   or with $Y_i=dummy$ (thanks
to the channelling constraints in line \ref{algo:channelling}). We
can build the tuple $I'$ on $X_1,..X_n, S,T$ where $X_i$ takes
value $v$, every $X_j$ with $j\in ub(S)$ and $I[Y_j]\in I[T]$
takes $I[Y_j]$, and the remaining $X_j$'s take any value in their
domain. $T$ is set to $I[T]$ plus the values taken by $X_j$'s with
$j\in lb(S)$. These values are in $ub(T)$ thanks to line
\ref{algo:Sbounds}. Finally, $S$ is set to $lb(S)$ plus the
indices of the $Y_j$'s with $I[Y_j]\in I[T]$. These indices are in
$ub(S)$ since the only $j$'s removed from $ub(S)$ in line
\ref{algo:Sbounds} are such that $D(Y_j)\cap ub(T)=\emptyset$,
which prevents $I[Y_j]$ from taking a value in $I[T]$.
%This construction ensures that $I'[T]=\bigcup_{j\in I'[S]}\{I'[X_j]\}$.
Thus $I'$ is a solution of \range\ with $I'[X_i]=v$. We have
proved that the $X_i$'s are hybrid consistent after \propagrange.

Suppose  a value $i\in ub(S)$ after line \ref{algo:channelling}.
Thanks to constraint in line \ref{algo:Sbounds} we know there
exists $v$ in $D(Y_i)\cap ub(T)$, and so, $v\in D(X_i)\cap ub(T)$.
Now, $X_i$ is hybrid consistent after line \ref{algo:channelling}.
Thus $X_i=v$  belongs to a solution of \range. If we modify this
solution by putting $i$ in $S$ and $v$ in $T$ (if not already
there), we keep a solution.

Completeness on $lb(S)$, $lb(T)$ and $ub(T)$ is proved in a
similar way.

\noindent\emph{Complexity}. The important thing to notice in
\propagrange\ is that constraints in lines
\ref{algo:occ}--\ref{algo:channelling} are propagated in
sequence. % and not all maintained BC,  GAC, or hybrid consistent.
Thus, \occurs\ is propagated only once, for a complexity in
$O(nd+n\cdot|lb(T)|^{3/2})$.
Lines \ref{algo:init}, \ref{algo:Sbounds}, and
\ref{algo:channelling} are in $O(nd)$.  Thus, the complexity of
\propagrange\ is
in $O(nd+n\cdot|lb(T)|^{3/2})$.
This reduces to
linear time complexity when $lb(T)$ is empty.

\noindent\emph{Incrementality}.
The overall complexity over a sequence of restrictions on $X_i$'s,
$S$ and $T$ is in   $O(n^2d^2)$. (See incrementality  of \occurs\
in Section \ref{sec:occurs}.)
\qed

Note that the Range constraint can be decomposed using the \gcc\
constraint. However, propagation on such a decomposition is in
$O(n^2d+n^{2.66})$ time complexity (see \cite{quietalCP04}).
\propagrange\ is thus significantly cheaper.

\section{Propagating the \roots\ constraint}
\label{sec:roots}

We now give a thorough theoretical analysis of the \roots\
constraint. In Section \ref{sec:roots:comp}, we provide a proof
%for the first time of the claim made in \cite{comicsIJCAI05}
that enforcing HC on \roots\ is NP-hard in general. Section
\ref{sec:roots:algo} presents a decomposition of the \roots\
constraint that permits us to propagate the \roots\ constraint
partially in linear time. Section \ref{sec:roots:poly} shows that
in many cases this decomposition does not destroy the global
nature of the \roots\ constraint as enforcing HC on the
decomposition achieves HC on the \roots\ constraint. Section
\ref{sec:roots:bc} shows that we can obtain BC on the \roots\
constraint by enforcing BC on its decomposition. Finally, we
provide some implementation details in Section \ref{sec:imp}.

%\subsection{Intractability of \roots\ Constraint}
\subsection{Complete propagation}
\label{sec:roots:comp}

Unfortunately, propagating the \roots\ constraint completely is
intractable in general. Whilst we made this claim in
\cite{comicsIJCAI05}, a proof has not yet been published. For this
reason, we give one here.

\begin{mytheorem}
Enforcing HC  on the \roots\ constraint is NP-hard.
\end{mytheorem}
\proof We transform \textsc{3Sat} into the problem of the
existence of a
solution %satisfying assignment
for \roots. Finding a hybrid support is thus NP-hard. Hence
enforcing HC on \roots\ is NP-hard. Let
$\varphi=\{c_1,\ldots,c_m\}$ be a 3CNF on the Boolean variables
$x_1,\ldots,x_n$. We build the constraint
$\roots([X_1,\ldots,X_{n+m}],S,T)$ as follows. Each Boolean
variable $x_i$ is represented by the variable $X_i$ with domain
$D(X_i)=\{i,-i\}$. Each clause $c_p=x_i\lor \neg x_j\lor x_k$ is
represented by the variable $X_{n+p}$ with  domain
$D(X_{n+p})=\{i,-j,k\}$. We build $S$ and $T$ in such a way that
it is impossible for both the index $i$ of a Boolean variable
$x_i$ and its complement $-i$ to belong to $T$. We set
$lb(T)=\emptyset$ and $ub(T)=\bigcup_{i=1}^n \{i,-i\}$, and
$lb(S)=ub(S)=\{n+1,\ldots,n+m\}$. An interpretation $M$ on the
Boolean variables $x_1,\ldots,x_n$ is a model of $\varphi$ iff the
tuple $\tau$ in which $\tau[X_i]=i$ iff $M[x_i]=0$ can be extended
to a solution of \roots. (This extension puts in $T$ value $i$ iff
$M[x_i]=1$ and assigns $X_{n+p}$ with the value corresponding to
the literal satisfying $c_p$ in $M$.)
\qed

We thus have to look for a lesser level of consistency for \roots\
or for particular cases on which HC is polynomial. We will show
that bound consistency is tractable and that, under conditions
often met in practice (e.g. one of the last two arguments of
\roots\ is ground), enforcing HC is also.

\subsection{A decomposition of \roots}
\label{sec:roots:algo}

To show that \roots\ can be propagated tractably, we will give a
straightforward decomposition into ternary constraints that can be
propagated in linear time. This decomposition does not destroy the
global nature of the \roots\ constraint since enforcing HC on the
decomposition will, in many cases, achieve HC on the original
\roots\ constraint, and since in all cases, enforcing BC on the
decomposition achieves BC on the original \roots\ constraint.
Given $\roots([X_1,\myldots,X_n], S, T)$, we decompose it into the
implications:
\begin{eqnarray*}
 i \in S & \rightarrow & X_i \in T \\
 X_i \in T & \rightarrow & i \in S
\end{eqnarray*}
where $i\in [1..n]$. We have to be careful how we implement such a
decomposition in a constraint solver. First, some solvers will not
achieve HC on such constraints (see Sec \ref{sec:imp} for more
details). Second, we need an efficient algorithm to be able to
propagate the decomposition in linear time. As we explain in more
detail in Sec \ref{sec:imp}, a constraint solver could easily take
quadratic time if
it is not incremental. % enough.

We first show that this decomposition prevents us from propagating
the \roots\ constraint completely. However, this is to be expected
as propagating \roots\ completely is NP-hard and this
decomposition is linear to propagate. In addition, as we later
show, in many circumstances met in practice, the decomposition
does not in fact hinder propagation.

\begin{mytheorem}
\label{hinderhc}
HC on $\roots([X_1,\myldots,X_n], S, T)$ is strictly stronger than
HC on $i \in S \rightarrow X_i \in T$, and $X_i \in T \rightarrow
i \in S$ for all $i \in [1..n]$.
\end{mytheorem}
\proof Consider $X_1 \in \{1,2\}$, $X_2 \in \{3,4\}$, $X_3 \in
\{1,3\}$, $X_4 \in \{2,3\}$, $lb(S)=ub(S)=\{3,4\}$,
$lb(T)=\emptyset$, and $ub(T)=\{1,2,3,4\}$. The decomposition is
HC. However, enforcing HC on \roots\  will prune 3 from $D(X_2)$.
\qed

In fact, enforcing HC on the decomposition achieves a level of
consistency between BC and HC on the original \roots\ constraint.
Consider $X_1 \in \{1,2,3\}$, $X_2 \in \{1,2,3\}$,
$lb(S)=ub(S)=\{1,2\}$, $lb(T)=\{\}$, and $ub(T)=\{1,3\}$. The
\roots\ constraint is BC. However, enforcing HC on the
decomposition  will remove 2 from the domains of $X_1$ and $X_2$. In
the next section, we identify exactly when the decomposition
achieves HC on \roots.

\subsection{Some special cases} % OF \roots}
\label{sec:roots:poly}

Many of the counting and occurrence constraints do not use the
\roots\ constraint in its more general form, but have some
restrictions on the variables $S$, $T$ or  $X_i$'s. For example,
it is often the case that $T$ or $S$ are ground. We select four
important cases that cover many of these uses of \roots\ and show
that enforcing HC on \roots\ is then tractable.
\begin{description}
\item[C1.] $\forall i\in lb(S), D(X_i)\subseteq lb(T)$
\item[C2.]  $\forall i\notin ub(S), D(X_i)\cap ub(T)=\emptyset$
\item[C3.]  $X_1, \myldots, X_n$ are ground
\item[C4.]  $T$ is ground
\end{description}
We will show that in any of these cases, we can achieve HC on
\roots\ simply by propagating the decomposition.
%Note that if $S$ is ground, enforcing
%HC on the decomposition will ensure $C1$ and $C2$.

\begin{mytheorem}\label{theo:hcroots}
If one of the conditions $C1$ to $C4$ holds, then enforcing HC on
$i \in S \rightarrow X_i \in T$, and $X_i \in T \rightarrow i \in
S$ for all $i \in [1..n]$ achieves HC on
$\roots([X_1,\myldots,X_n], S, T)$.
\end{mytheorem}
\proof
Our proof will exploit the
following properties that are guaranteed to hold when we have
enforced HC on the decomposition.

\newcommand{\pub}{\texttt{Pout\_ubT}}
\newcommand{\plb}{\texttt{Pin\_lbT}}
\newcommand{\cub}{\texttt{Cout\_ubT}}
\newcommand{\clb}{\texttt{Cin\_lbT}}
\newcommand{\lub}{\texttt{L\_ubT}}
\newcommand{\llb}{\texttt{L\_lbT}}

\begin{description}
\item[P1] if $D(X_i)\subseteq lb(T)$    then  $i\in lb(S)$
\item[P2] if $D(X_i)\cap ub(T) =\emptyset$ then  $i\notin ub(S)$
\item[P3] if $i\in lb(S)$ then  $D(X_i)\subseteq ub(T)$
\item[P4] if $i\notin ub(S)$ then $D(X_i)\cap lb(T)=\emptyset$
\item[P5] if $D(X_i)=\{v\}$ and $i\in lb(S)$
  then $v\in lb(T)$
\item[P6] if $D(X_i)=\{v\}$  and  $i\notin
  ub(S)$ then $v\notin ub(T)$
\item[P7] if $i$ is added to $lb(S)$ by the constraint $X_i\in T\to
  i\in S$ then $D(X_i)\subseteq lb(T)$
\item[P8] if $i$ is deleted from $ub(S)$ by the constraint $i\in S \to X_i\in T$
  then
  $D(X_i)\cap ub(T)=\emptyset$
\end{description}

\emph{Soundness.} Immediate.

\emph{Completeness.} We assume that  one of the conditions C1---C4
holds and the decomposition is HC. We will first prove that  the
\roots\ constraint is satisfiable. Then, we will prove that, for
any $X_i$, all the values in $D(X_i)$ belong to a solution of
\roots, and that the bounds on $S$ and $T$ are as tight as
possible.

We  prove that  the
\roots\ constraint is satisfiable.
Suppose that  one of the conditions C1---C4 holds and  that the
decomposition is   HC.
Build the following tuple $\tau$ of values for the $X_i$, $S$, and
$T$. Initialise $\tau[S]$ and $\tau[T]$ with $lb(S)$ and $lb(T)$
respectively.  Now, let us consider the four conditions separately.

(C1) For each $i\in \tau[S]$,
  choose any value $v$ in $D(X_i)$ for
  $\tau[X_i]$. From   the assumption and from
  property P7 we deduce that $v$ is in $lb(T)$, and so in $\tau[T]$. For
  each other $i$, assign $X_i$ with any value in $D(X_i)\setminus
  lb(T)$. (This set is not empty thanks to property P1.) $\tau$
  obviously satisfies \roots.

(C2) For each $i\in\tau[S]$,  choose any value in
  $D(X_i)$  for $\tau[X_i]$. By construction such a value is in
  $ub(T)$ thanks to property P3.  If necessary, add $\tau[X_i]$ to
  $\tau[T]$. For each other $i\in ub(S)$,  assign $X_i$ with any value in
  $D(X_i)\setminus \tau[T]$ if possible. Otherwise assign $X_i$ with
  any value in $D(X_i)$ and add $i$ to $\tau[S]$. For each $i\notin
  ub(S)$, assign $X_i$ any value from its domain. By assumption
% we know that $i$ has been removed during the propagation of
% \roots,
and by property P8 we know that $D(X_i)\cap
  ub(T)=\emptyset$.  Thus, $\tau$
  satisfies \roots.

(C3) $\tau[X_i]$ is already assigned for all $X_i$.
  For each $i\in \tau[S]$, property P5 tells us that
  $\tau[X_i]$ is in  $\tau[T]$, and for  each $i\notin lb(S)$,
  property P1 tells us that $\tau[X_i]$ is outside $lb(T)$.
  $\tau$ satisfies \roots.

(C4) For each $i\in \tau[S]$  choose any value $v$ in $D(X_i)$ for
  $\tau[X_i]$. Property P3 tells us $v\in ub(T)$. By assumption, $v$
  is thus in $\tau[T]$.  For
  each $i$ outside $ub(S)$, assign $X_i$ with any value $v$ in
  $D(X_i)$. ($v$ is outside $\tau[T]$ by assumption and property P4). For
  each other $i$, assign $X_i$ with any value in $D(X_i)$ and update
  $\tau[S]$ if necessary. $\tau$ satisfies \roots.

We have proved that the \roots\ constraint has a solution.
We now  prove that for  any value in $ub(S)$ or in $ub(T)$ or in
$D(X_i)$  for any $X_i$,  we can transform the  arbitrary
solution of \roots\ into a solution that contains that value.
Similarly, for any value not in  $lb(S)$ or not in
$lb(T)$, we can transform the  arbitrary
solution of \roots\ into a solution that does not contain  that  value.

Let us prove that  $lb(T)$ is tight. Suppose the tuple $\tau$ is a
solution of the \roots\ constraint. Let $v\not\in lb(T)$ and $v\in
\tau[T]$. We show that there exists a solution with
$v\not\in\tau[T]$. (Remark that this case is irrelevant to
condition C4.) We remove $v$ from $\tau[T]$. For each $i\not\in
lb(S)$ such that $\tau[X_i]=v$ we remove $i$ from $\tau[S]$. With
C1 we are sure that none of the $i$ in $lb(S)$ have $\tau[X_i]=v$,
thanks to property P7 and the fact that $v\not\in lb(T)$.  With C3
we are sure that none of the $i$ in $lb(S)$ have $\tau[X_i]=v$,
thanks to property P5 and the fact that $v\not\in lb(T)$. There
remains to check C2. For each $i\in lb(S)$, we know that  $\exists
v'\neq v, v'\in D(X_i)\cap ub(T)$, thanks to properties P3 and P5.
We set $X_i$ to $v'$ in $\tau$, we add $v'$ to $\tau[T]$  and add
all $k$ with $\tau[X_k]=v'$ to $\tau[S]$. We are sure that $k\in
ub(S)$ because $v'\in ub(T)$ plus  condition C2 and  property P8.

Completeness on $ub(T)$,  $lb(S)$, $ub(S)$ and  $X_i$'s are shown
with  similar proofs.
Let $v\in ub(T)\setminus \tau[T]$. (Again C4
is
 irrelevant.) We show that there exists a solution with $v\in\tau[T]$.
 Add $v$ to $\tau[T]$ and for each $i\in ub(S)$, if
 $\tau[X_i]=v$, put $i$ in $\tau[S]$. C2 is solved thanks to
 property P8 and the fact that $v\in
 ub(T)$. C3 is solved thanks to
 property P6 and the fact that $v\in
 ub(T)$.
 There remains to check
 C1.  For each $i\not\in ub(S)$ and $\tau[X_i]=v$, we know
 that  $\exists v'\neq v, v'\in D(X_i)\setminus lb(T)$
 (thanks to properties P4 and P6).
 We set $X_i$ to $v'$ in $\tau$ and remove
 $v'$ from $\tau[T]$. Each $k$ with $\tau[X_k]=v'$ is removed from
 $\tau[S]$, and this is possible because we are in condition C1,
 $v'\not\in lb(T)$, and thanks to property P7.

 Let $v\in D(X_i)$ and $\tau[X_i]=v', v' \neq v$. (C3 is
 irrelevant.)
 Assign $v$ to $X_i$
 in $\tau$. If both $v$ and $v'$ or none of them are in $\tau[T]$, we
 are done. There remain two cases.
 First, if $v\in \tau[T]$ and
 $v'\not\in \tau[T]$, the two alternatives to satisfy \textsc{Roots}
 are to add $i$ in $\tau[S]$ or to remove $v$ from $\tau[T]$.
 If $i\in ub(S)$, we add $i$ to $\tau[S]$ and we are done. If $i\not\in
 ub(S)$, we know that $v\not\in lb(T)$
 thanks to property P4. So, $v$ is removed from $\tau[T]$ and we are
 sure that the $X_j$'s can be updated consistently for the same reason
 as in the proof of $lb(T)$.
 Second, if $v\not\in \tau[T]$ and
 $v'\in \tau[T]$, the two alternatives to satisfy \textsc{Roots}
 are to remove $i$ from $\tau[S]$ or to add $v$ to $\tau[T]$.
 If $i\notin lb(S)$, we remove $i$ from $\tau[S]$ and we are done. If
 $i\in lb(S)$, we know that $v\in ub(T)$
 thanks to property P3. So, $v$ is added to $\tau[T]$ and we are
 sure that the $X_j$'s can be updated consistently for the same reason
 as in the proof of $ub(T)\setminus \tau[T]$.

 Let $i\not\in lb(S)$ and $i\in \tau[S]$. We show that there exists a
 solution with $i\not\in\tau[S]$.  We remove $i$ from $\tau[S]$. Thanks
 to property P1, we know that $D(X_i)\not\subseteq lb(T)$. So, we set
 $X_i$ to a value $v'\in D(X_i)\setminus lb(T)$. With C4 we
 are done because we are sure $v'\not\in \tau[T]$. With conditions C1,
 C2, and C3, if $v'\in \tau[T]$,
 we remove it from $\tau[T]$ and we are
 sure that the $X_j$'s can be updated consistently for the same reason
 as in the proof of $lb(T)$.

 Let $i\in ub(S)\setminus \tau[S]$.
 We show that there exists a solution with $i\in\tau[S]$.
 We add $i$ to $\tau[S]$. Thanks
 to property P2, we know that
 $D(X_i)\cap ub(T)\neq\emptyset$. So, we set
 $X_i$ to a value $v'\in D(X_i)\cap ub(T)$. With condition C4 we
 are done because we are sure $v'\in \tau[T]$. With conditions C1, C2,
 and C3, if $v'\not\in \tau[T]$,
 we add it to $\tau[T]$ and we are
 sure that the $X_j$'s can be updated consistently for the same reason
 as in the proof of $ub(T)\setminus \tau[T]$. \qed

\subsection{Bound consistency} % ON \roots}
\label{sec:roots:bc} In addition to being able to enforce HC on
\roots\ in some special cases, enforcing HC on the decomposition
always enforces a level of consistency at least as strong as BC.
In fact, in any situation (even those where enforcing HC is
intractable), enforcing BC on the
decomposition enforces BC on the \roots\ constraint. % in linear time.

\begin{mytheorem}
Enforcing BC on $i \in S \rightarrow X_i \in T$, and $X_i \in T
\rightarrow i \in S$ for all $i \in [1..n]$ achieves BC on
$\roots([X_1,\myldots,X_n], S, T)$.
\end{mytheorem}
\proof
\emph{Soundness.} Immediate. \\
\emph{Completeness.} The proof follows the same structure as that
in Theorem \ref{theo:hcroots}. We relax the properties P1--P4 into
properties P1'--P4'.
\begin{description}
\item[P1'] if $[min(X_i),max(X_i)]\subseteq lb(T)$    then  $i\in lb(S)$
\item[P2'] if $[min(X_i),max(X_i)]\cap ub(T)=\emptyset$
  then  $i\not\in ub(S)$
\item[P3']if $i\in lb(S)$ then  the bounds of $X_i$ are included in $ub(T)$
\item[P4'] if $i\notin ub(S)$ then the bounds of $X_i$ are outside $lb(T)$
\end{description}

Let us prove that  $lb(T)$ and $ub(T)$ are tight. Let $o$ be the
total ordering on $D=\bigcup_i D(X_i)\cup ub(T)$. Build the tuples
$\sigma$ and $\tau$ as follows:
For each $v\in lb(T)$: put $v$ in $\sigma[T]$ and $\tau[T]$.
%\item
For each $v\in ub(T)\setminus lb(T)$, following $o$, do: put $v$
in $\sigma[T]$ or  $\tau[T]$ alternately.
%\item
For each $i \in lb(S)$, P3' guarantees that both $min(X_i)$ and
$max(X_i)$ are in $ub(T)$. By construction of $\sigma[T]$ (and
$\tau[T]$) with alternation of values, if $min(X_i)\neq max(X_i)$,
we are sure that there exists  a value in $\sigma[T]$ (in
$\tau[T]$) between $min(X_i)$ and $max(X_i)$. In the case
$|D(X_i)|=1$, P5  guarantees  that the only value is in
$\sigma[T]$ (in $\tau[T]$). Thus, we assign $X_i$ in $\sigma$ (in
$\tau$) with such a value in $\sigma[T]$ (in $\tau[T]$).
%\item
For each $i\notin ub(S)$, we assign $X_i$ in $\sigma$ with a value
in $[min(X_i), $ $max(X_i)]\setminus \sigma[T]$ (the same for
$\tau$). We know  that such a value exists with the same reasoning
as for $i\in lb(S)$ on alternation of values, and thanks to P4'
and P6.
%\item
We complete $\sigma$ and $\tau$ by building $\sigma[S]$ and
$\tau[S]$ consistently with the assignments of $X_i$ and~$T$.
%\end{itemize}
The resulting tuples satisfy \roots. From this we  deduce that
$lb(T)$ and $ub(T)$ are BC as all values in $ub(T)\setminus lb(T)$
are either in $\sigma$ or in $\tau$, but not both.

We  show that the $X_i$ are BC. Take any $X_i$ and its lower bound
$min(X_i)$. If $i\in lb(S)$ we know that $min(X_i)$ is in $T$
either in $\sigma$ or in $\tau$ thanks to P3' and by construction
of $\sigma$ and $\tau$. We assign $min(X_i)$ to $X_i$ in the
relevant tuple. This remains a solution of \roots. If $i\notin
ub(S)$, we know that $min(X_i)$ is outside $T$ either in $\sigma$
or in $\tau$ thanks to P4' and by construction of $\sigma$ and
$\tau$. We assign $min(X_i)$ to $X_i$ in the relevant tuple. This
remains a solution of \roots. If $i\in ub(S)\setminus lb(S)$,
assign $X_i$ to $min(X_i)$ in $\sigma$. If
$min(X_i)\notin\sigma[T]$, remove $i$ from $\sigma[S]$ else add
$i$ to  $\sigma[S]$. The tuple obtained is a solution of \roots\
using the lower bound of $X_i$. By the same reasoning, we show
that the upper bound of $X_i$ is BC also, and therefore, all
$X_i$'s are BC.

We prove that $lb(S)$ and $ub(S)$ are BC  with  similar proofs.
 Let us show that $ub(S)$ is BC. Take any $X_i$ with $i\in ub(S)$ and
 $i\notin\sigma[S]$. Since $X_i$ was assigned any value from
 $[min(X_i),max(X_i)]$ when $\sigma$ was built, and since we know
 that $[min(X_i),max(X_i)]\cap ub(T)\neq\emptyset$ thanks to P2',
 we can modify $\sigma$ by assigning $X_i$ a value in $ub(T)$,
 putting the value in $T$ if not already there, and adding $i$ into
 $S$. The tuple obtained satisfies \textsc{Roots}. So $ub(S)$ is
 BC.

 There remains to show that $lb(S)$ is BC. Thanks to P1', we know
 that values $i\in ub(S)\setminus lb(S)$ are such that
 $[min(X_i),max(X_i)]\setminus lb(T)\neq\emptyset$. Take
 $v\in [min(X_i),max(X_i)]\setminus lb(T)$. Thus, either $\sigma$
 or $\tau$ is such that $v\notin T$. Take the corresponding tuple,
 assign $X_i$ to $v$ and remove $i$ from $S$. The modified tuple is
 still a solution of \textsc{Roots} and $lb(S)$ is BC. \qed

\subsection{Implementation details} \label{sec:imp}

This decomposition of the \roots\ constraint can be implemented in
many solvers using disjunctions of membership and
negated membership constraints: ${\tt
or}({\tt member}(i,S),{\tt not\-member}(X_i,T))$ and ${\tt
or}({\tt not\-member}(i,S),{\tt member}(X_i,T))$. However, this
requires a little care. Unfortunately, some existing solvers (like
Ilog Solver) may not achieve HC on such disjunctions of
primitives. For instance,  the
negated membership constraint %$X_i\in T$
${\tt notmember}(X_i,T)$ may be activated only if
%is set to false only if
$X_i$ is instantiated with a value of $T$ (whereas it should be as
soon as
$D(X_i)\subseteq lb(T)$% or $D(X_i)\cap ub(T)=\emptyset$
). We have to ensure that the solver wakes up when it should to
ensure we achieve HC. As we explain in the complexity proof, we
also have to be careful that the solver does not wake up too often
or we will lose the optimal $O(nd)$ time complexity which can be
achieved.

\begin{mytheorem}\label{theo:algo1}
It is possible to enforce HC (or BC) on the decomposition of
$\roots([X_1,\myldots,X_n], S, T)$ in $O(nd)$ time, where
$d=max(\forall i.|D(X_i)|,|ub(T)|)$.
\end{mytheorem}
\proof The decomposition of \roots\ is composed of $2n$
constraints. To obtain an overall complexity in $O(nd)$, the total
amount of work spent propagating each of these constraints must be
in $O(d)$ time.

First, it is necessary that each of the $2n$ constraints of the
decomposition is not called for propagation more than $d$ times.
Since $S$ can be modified up to $n$ times ($n$ can be larger than
$d$) it is important that not all constraints are called for
propagation at each change in $lb(S)$ or $ub(S)$. By implementing
'propagating events' as described in \cite{choco00,schstuCP04}, we
can ensure that when a value $i$ is added to $lb(S)$ or removed
from $ub(S)$, constraints $j\in S\to X_j\in T$ and  $X_j\in T\to
j\in S$, $j\neq i$, are not called for propagation.

Second, we show that  enforcing HC on constraint $i\in S \to
X_i\in T$ is in $O(d)$ time. Testing the precondition (does $i$
belong to $lb(S)$?) is constant time. If true, removing from
$D(X_i)$ all values not in $ub(T)$ is in $O(d)$ time and updating
$lb(T)$ (if $|D(X_i)|=1$) is constant time. Testing that the
postcondition is false (is $D(X_i)$ disjoint from $ub(T)$?)  is in
$O(d)$ time. If false, updating $ub(S)$ is constant time. Thus HC
on $i\in S \to X_i\in T$ is in $O(d)$ time. Enforcing HC on
$X_i\in T\to i\in S$ is in $O(d)$ time as well because testing the
precondition ($D(X_i)\subseteq lb(T)$?) is in $O(d)$ time,
updating $lb(S)$ is constant time,  testing that the postcondition
is false ($i\notin ub(S)$?) is constant time, and removing from
$D(X_i)$ all values in $lb(T)$ is in $O(d)$ time and updating
$ub(T)$ (if $|D(X_i)|=1$) is constant time.

When $T$ is modified, all constraints are potentially concerned.
Since $T$ can be modified up to $d$ times, we can have $d$ calls
of the propagation in $O(d)$ time for each
%a brute force propagation would
%lead to $d$ propagations
of the $2n$ constraints. It is thus important that the propagation
of the $2n$ constraints is \emph{incremental} to avoid an
$O(nd^2)$ overall complexity. An algorithm for $i\in S \to X_i\in
T$ is incremental if the complexity  of calling the propagation of
the constraint $i\in S \to X_i\in T$ up to $d$ times (once for
each change in $T$ or $D(X_i)$)  is the same as propagating the
constraint once. This can be achieved by an AC2001-like algorithm
that stores the last value found in $D(X_i)\cap ub(T)$, which is a
witness that the postcondition can be true. (Similarly,  the last
value found in $D(X_i)\setminus lb(T)$ is a witness that the
precondition of the constraint $X_i\in T\to i\in S$ can be false.)
Finally, each time $lb(T)$ (resp. $ub(T)$) is  modified, $D(X_i)$
must be updated for each $i$ outside $ub(S)$ (resp. inside
$lb(S)$). If the propagation mechanism of the solver  provides the
values that have been added to $lb(T)$ or removed from $ub(T)$ to
the propagator of the $2n$ constraints (as described in
\cite{vanhetal92}),  updating a given $D(X_i)$ has a total
complexity  in $O(d)$ time for the $d$ possible changes in $T$.
The proof that BC can also be enforced in linear time follows a
similar argument. \qed

\section{A catalog of decompositions using \range\ and \roots}
\label{sec:catalog}

We have shown how to propagate the \range\ and \roots\
constraints. Specification of counting and occurrence
constraints using \range\ and \roots\ will thus be  executable.
\range\ and \roots\
permit us to decompose counting and occurrence global constraints into
more primitive
constraints, each of which having  an associated polynomial
propagation algorithm. In some cases, such  decomposition does not
hinder propagation. In other cases, enforcing local consistency on
the global constraint is intractable, and decomposition is one
method to obtain a polynomial propagation algorithm
\cite{besetalAAAI04,besetalCP04,besetalConstraints07}.

In a technical report \cite{range-roots-tech-report}, we present a
catalog containing over 70 global constraints from
\cite{beldiceanu3} specified with the help of the \range\ and
\roots\ constraints.
Here we %have room
%to
present
%just a dozen
a few of the more important constraints. In  the subsequent five
subsections, we list some counting and occurrence  constraints which can be
specified using \range\ constraints, using \roots\ constraints,
and using both \range\ and \roots\ constraints. 
We also show that \range\ and \roots\ can  be used to specify 
\emph{open} global constraints, a  new kind of global constraints introduced
recently.  We finally 
include problem domains other than counting and occurrence to 
illustrate the wide range of global constraints expressible in
terms of \range\ and \roots.

\subsection{Applications of \range\ constraint}

\range\ constraints are often useful to specify constraints on the
values used by a sequence of variables.

\subsubsection{All different}
\label{app:alldiff}

The \alldifferent\ constraint forces a sequence of variables to
take different values from each other. Such a constraint is useful
in a wide range of problems (e.g. allocation of activities to
different slots in a time-tabling problem).
It  can be propagated efficiently \cite{regin1}. It can also be
decomposed with a single \range\ constraint:

%\vspace{-0.7em}
\begin{eqnarray*}
& & \alldifferent([X_1,\myldots,X_n])  \ \ \myiff    \\
& & \range([X_1,\myldots,X_n],\{1,\myldots,n\},T) \ \And \ |T|=n
\end{eqnarray*}
%\vspace{-1em}

A special but nevertheless important case of this constraint is
the \permutation\ constraint. This is an \alldifferent\ constraint
where we additionally know $R$, the set of values to be taken.
That is, the sequence of variables is a permutation of the values
in $R$ where $|R|=n$. This also can be decomposed using a single
\range\ constraint:

%\vspace{-0.7em}
\begin{eqnarray*}
& & \permutation([X_1,\myldots,X_n],R) \ \ \myiff \\
& & \range([X_1,\myldots,X_n],\{1,\myldots,n\},R)
\end{eqnarray*}
%\vspace{-1em}

Such a decomposition of the
\permutation\ constraint %into a \range\ constraint
obviously does not hinder propagation. However, decomposition of
\alldifferent\ into
a \range\ %and cardinality
constraint does.
This example illustrates that, whilst many global constraints can be
expressed in terms of \range\ and \roots, there are some global
constraints like \alldifferent\ for which it is worth developing
%we need to develop
specialised propagation algorithms. Nevertheless, \range\ and
\roots\ provide a means of propagation for such constraints in the
absence of specialised algorithms. They can also enhance the
existing propagators. For instance, HC on the \range\ decomposition
is incomparable to AC on the decomposition of \alldifferent\ which
uses a clique of binary inequality constraints. Thus, we may be able
to obtain more pruning by using both decompositions.
\begin{mytheorem}\label{theorem:alldiff}
(1) GAC on \permutation\ is equivalent to HC on the decomposition with
\range.
(2) GAC on \alldifferent\ is stronger than HC on the decomposition
with \range.
(3) AC on the decomposition of \alldifferent\ into binary
inequalities is incomparable to HC on the decomposition with
\range.
\end{mytheorem}
\myproof (1) \permutation\ can be encoded as a single \range.
Moreover, since $R$ is fixed, HC is equivalent to AC. (2) Consider
$X_1$, $X_2 \in \{1,2\}$, $X_3 \in \{1,2,3,4\}$, and $\{1,2\}
\subseteq T \subseteq \{1,2,3,4\}$. Then
$\range([X_1,X_2,X_3],\{1,2,3\},T)$ and
$|T|=3$ are both HC, but %enforcing GAC on
$\alldifferent([X_1,X_2,X_3])$ is not GAC.
%prunes 1 and 2 from $X_3$.
(3) Consider $X_1$, $X_2 \in \{1,2\}$, $X_3 \in \{1,2,3\}$, and
$T=\{1,2,3\}$. Then $X_1 \neq X_2$, $X_1 \neq X_3$ and $X_2 \neq
X_3$ are AC but $\range([X_1,X_2,X_3],\{1,2,3\},T)$ is not HC.
Consider $X_1$, $X_2 \in \{1,2,3,4\}$, $X_3 \in \{2\}$, and $\{2\}
\subseteq T \subseteq \{1,2,3,4\}$. Then
$\range([X_1,X_2,X_3],\{1,2,3\},T)$ and $|T|=3$ are HC. But $X_1
\neq X_3$ and $X_2 \neq X_3$ are not AC.
%The other result holds immediately.
\myqed

\subsubsection{Disjoint}

We may require that two sequences of variables be disjoint (i.e.
have no value in common). For instance, two sequences of tasks
sharing the same resource might be required to be disjoint in
time. The $\disjoint([X_1,\myldots,X_n],[Y_1,\myldots,Y_m])$
constraint introduced in \cite{beldiceanu3} ensures $X_i \neq Y_j$
for any $i$ and $j$. We prove here that we cannot expect to
enforce GAC on such a constraint as it is NP-hard to do so in
general.
\begin{mytheorem}
Enforcing GAC on %$\disjoint([X_1,\myldots,X_n],[Y_1,\myldots,Y_m])$ is
\disjoint\ is NP-hard.
\end{mytheorem}
\myproof We reduce 3-SAT to the problem of deciding if a
\disjoint\ constraint has any satisfying assignment. Finding
support is therefore NP-hard. Consider a formula $\varphi$ with
$n$ variables and $m$ clauses. For each Boolean variable $x$, we
let $X_x \in \{ x, \neg x\}$ and $Y_j \in \{ x, \neg y,z\}$ where
the $j$th clause in $\varphi$ is $x \vee \neg y \vee z$. If
$\varphi$ has a model then the \disjoint\ constraint has a
satisfying assignment in which the $X_x$ take the literals false
in this model. \myqed

One way to propagate a \disjoint\
constraint is to decompose it into two
\range\ constraints:
\begin{eqnarray*}
& & \disjoint([X_1,\myldots,X_n],[Y_1,\myldots,Y_m])  \ \ \myiff    \\
& & \range([X_1,\myldots,X_n],\{1,\myldots,n\},S) \ \And \ \\
& & \range([Y_1,\myldots,Y_m],\{1,\myldots,m\},T) \ \And \ S \cap
T=\{\}
\end{eqnarray*}
Enforcing HC on this decomposition is
polynomial.
Decomposition thus offers a simple and promising
method to propagate a \disjoint\ constraint.
Not surprisingly, the
decomposition hinders propagation
(otherwise we would have a polynomial algorithm
for a NP-hard problem).
\begin{mytheorem}
GAC on %$\disjoint([X_1,\myldots,X_n],[Y_1,\myldots,Y_m])$
\disjoint\ is stronger than
HC on the decomposition.
\end{mytheorem}
\myproof
Consider $X_1, Y_1 \in \{1,2\}$,
$X_2, Y_2 \in \{1,3\}$,
$Y_3 \in \{2,3\}$ and
$\{\} \subseteq S, T \subseteq \{1,2,3\}$.
Then
$\range([X_1,X_2],\{1,2\},S)$
and $\range([Y_1,Y_2,Y_3],\{1,2,3\},T)$ are HC,
and $S \cap T=\{\}$ is BC.
However, enforcing GAC on $\disjoint([X_1,X_2],[Y_1,Y_2,Y_3])$
prunes 3 from $X_2$ and 1 from both $Y_1$ and $Y_2$.
\myqed

\subsubsection{Number of values}

The \nvalue\ constraint is useful in a wide range of problems
involving resources since it counts the number of distinct values
used by a sequence of variables 
\cite{pachet1,comicsNVALUE,comicsCONSTRAINTS06}. As we saw in Section
\ref{sec:lang},
$\nvalue([X_1,\myldots,X_n],N)$ holds iff $N=|\{ X_i \ | \ 1 \leq
i \leq n\}|$. The \alldifferent\ constraint is a special case of
the \nvalue\ constraint in which $N=n$. Unfortunately, it is
NP-hard in general to enforce GAC on a \nvalue\ constraint
\cite{besetalAAAI04}. However, there is an $O(n \log(n))$ algorithm
%based on interval graphs
to enforce a level of consistency similar to BC \cite{bcp01}. An
alternative and even simpler way to implement this constraint is
with a \range\ constraint:

%\vspace{-0.7em}
\begin{eqnarray*}
& & \nvalue([X_1,\myldots,X_n],N)  \ \ \myiff    \\
& & \range([X_1,\myldots,X_n],\{1,\myldots,n\},T) \ \And \ |T|=N
\end{eqnarray*}
%\vspace{-1em}

HC on this decomposition is incomparable to BC on the \nvalue\
constraint.
\begin{mytheorem}
BC on %$\nvalue([X_1,\myldots,X_n],N)$
\nvalue\ is incomparable to HC on the decomposition.
%$\range([X_1,\myldots,X_n],\{1,\myldots,n\},T)$
%and on $|T|=N$.
\end{mytheorem}
\myproof Consider $X_1, X_2 \in \{1,2\}$, $X_3 \in \{1,2,3,4\}$,
$N \in \{3\}$ and $\{\} \subseteq T \subseteq \{1,2,3,4\}$. Then
$\range([X_1,X_2,X_3],\{1,2,3\},T)$ and $|T|=N$ are both HC.
However, enforcing BC on $\nvalue([X_1,X_2,X_3],N)$ prunes 1 and 2
from $X_3$.

Consider $X_1,X_2,X_3 \in \{1,3\}$ and $N \in \{3\}$. Then
$\nvalue([X_1,X_2,X_3],N)$ is BC. However, enforcing HC on
$\range([X_1,X_2,X_3],\{1,2,3\},T)$ makes $\{\} \subseteq T
\subseteq \{1,3\}$ which will cause $|T|=3$ to fail. \myqed

\subsubsection{Uses} \label{uses}

In \cite{beldiceanu7},  propagation algorithms achieving GAC and
BC are proposed for the  \usedby\ constraint.
\usedby$([X_1,\myldots,X_n],[Y_1,\myldots,Y_m])$ holds iff the
\emph{multiset} of values assigned to $Y_1,\myldots,Y_m$ is a
subset of the \emph{multiset} of values assigned to
$X_1,\myldots,X_n$.
We now introduce a %generalization : Toby, some readers may
% not agree it is a generalization but a specialization!
variant of the \usedby\ constraint called the \uses\ constraint.
\uses$([X_1,\myldots,X_n],[Y_1,\myldots,Y_m])$ holds iff the \emph{set} of
values assigned to $Y_1,\myldots,Y_m$ is a subset of the
\emph{set} of values assigned to $X_1,\myldots,X_n$. That is,
\usedby\ takes into account the number of times  a value is used
while \uses\ does not. Unlike the \usedby\ constraint,
%we prove that we cannot expect to
%enforce GAC on such a constraint in polynomial time as it is NP-hard to do so in
%%%%%general.
enforcing GAC on \uses\ is NP-hard. % in general.
\begin{mytheorem}
Enforcing GAC on \uses\ is NP-hard.
\end{mytheorem}
\myproof We reduce 3-SAT to the problem of deciding if a \uses\
constraint has a solution.
%satisfying assignment.
Finding support is therefore NP-hard. Consider a formula $\varphi$
with $n$ Boolean variables and $m$ clauses. For each Boolean
variable $x$, we introduce a variable $X_x \in \{ x, -x\}$. For
each clause $c_j=x \vee \neg y \vee z$, we introduce $Y_j \in \{
x,-y,z\}$. Then $\varphi$ has a model iff the \uses\ constraint
has a satisfying assignment, and $x$ is true iff  $X_x=x$. \myqed

One way to propagate a \uses\ constraint is to decompose it using
\range\  constraints:

%\vspace{-0.7em}
\begin{eqnarray*}
& & \uses([X_1,\myldots,X_n],[Y_1,\myldots,Y_m])  \ \ \myiff    \\
& & \range([X_1,\myldots,X_n],\{1,\myldots,n\},T) \ \And \\
& & \range([Y_1,\myldots,Y_m],\{1,\myldots,m\},T') \ \And \ T'
\subseteq T
\end{eqnarray*}
%\vspace{-1em}

Enforcing HC on this decomposition is polynomial. Not
surprisingly,
this % decomposition
hinders propagation (otherwise we would have a polynomial
algorithm for a NP-hard problem).
\begin{mytheorem}
GAC on \uses\ is stronger than HC on the decomposition.
\end{mytheorem}
\myproof Consider $X_1 \in \{1,2,3,4\}$, $X_2 \in \{1,2,3,5\}$,
$X_3,X_4 \in \{4,5,6\}$, $Y_1 \in \{1,2\}$, $Y_2 \in \{1,3\}$, and
$Y_3 \in \{2,3\}$. The decomposition is HC while GAC on \uses\
prunes 4 from the domain of $X_1$ and 5 from the domain of
$X_2$.\myqed

Thus, decomposition is a simple method to obtain a polynomial
propagation algorithm.
% as achieving GAC is
%intractable.

\subsection{Applications of \roots\ constraint}

\range\ constraints are often useful to specify constraints on the
values used by a sequence of variables. \roots\ constraint, on the
other hand, are useful to specify constraints on the variables
taking particular values.

\subsubsection{Global cardinality}
\label{sec:gcc}

The global cardinality constraint introduced in \cite{regin2}
constrains the number of times values are used. We consider a
generalization in which the number of occurrences of a value may
itself be an integer variable. More precisely, 
$\gcc([X_1,\myldots,X_n],[d_1,\myldots,d_m],[O_1,\myldots,O_m])$
holds iff $|\{ i \ | \ X_i = d_j \}|=O_j$ for all $j$. Such a
\gcc\ constraint can be decomposed into
%a set of \among\ constraints
% (and thus into
a set of \roots\ constraints: % ):

%\vspace{-0.7em}
\begin{eqnarray*}
& & \gcc([X_1,\myldots,X_n],[d_1,\myldots,d_m],[O_1,\myldots,O_m])  \ \ \myiff    \\
& & \forall i \ . \ \roots([X_1,\myldots,X_n],S_i,\{d_i\}) \ \And
\ |S_i| = O_i
\end{eqnarray*}
%\vspace{-1em}

Enforcing HC on these \roots\ constraints is polynomial since the
sets $\{d_i\}$ are ground (See Theorem~\ref{theo:hcroots}).
Enforcing GAC on a generalised \gcc\ constraint is NP-hard, but we
can enforce GAC on the $X_i$ and BC on the $O_j$ in polynomial time
using a specialised algorithm \cite{quietalCP04}. This is more than
is achieved by the decomposition.

\begin{mytheorem}
GAC on the $X_i$ and BC on the $O_j$ of a  \gcc\ constraint is
stronger than HC on the decomposition using \roots\ constraints.
\end{mytheorem}
\myproof As sets are represented by their bounds, HC on the
decomposition cannot prune more on the $O_j$ than BC does on the
\gcc. To show strictness, consider $X_1, X_2 \in \{1,2\}$, $X_3
\in \{1,2,3\}$, $d_i=i$ and $O_1, O_2, O_3 \in \{0,1\}$.
% Each of the \roots\
% constraints is HC (with $\{\} \subseteq S_1, S_2 \subseteq
% \{1,2,3\}$ and $\{\} \subseteq S_3 \subseteq \{3\}$) and each of
% the cardinality constraints is BC.
The decomposition is HC (with $\{\} \subseteq S_1, S_2 \subseteq
\{1,2,3\}$ and $\{\} \subseteq S_3 \subseteq \{3\}$). However,
enforcing GAC on the $X_i$
and BC on the $O_j$ of % BC (and thus GAC) on
the \gcc\ constraint will prune 1 and 2 from $X_3$ and 0 from
$O_1$, $O_2$ and $O_3$. \myqed

This illustrates another global constraint for which it is
worth developing a
%we need to develop
specialised propagation algorithm.

\subsubsection{Among}\label{sec:among}

The \among\ constraint was introduced in CHIP to help model
resource allocation problems like car sequencing
\cite{beldiceanu5}. It counts the number of variables using values
from a given set. $\among([X_1,\myldots,X_n],$
$[d_1,\myldots,d_m],N)$
holds iff $N=|\{ i \ | \ X_i \in\{d_1,\myldots,d_m\}\}|$.

An alternative way to propagate the \among\ constraint is to
decompose it using a \roots\ constraint:

%\vspace{-0.7em}
\begin{eqnarray*}
& & \among([X_1,\myldots,X_n],[d_1,\myldots,d_m],N)  \ \ \myiff    \\
& & \roots([X_1,\myldots,X_n],S,\{d_1,\myldots,d_m\}) \ \And \
|S|=N
\end{eqnarray*}
%\vspace{-1em}

It is polynomial %Even though it is NP-hard
to enforce HC on this case of the \roots\ constraint
since the target set is ground. This decomposition also does not
hinder propagation. It is therefore a potentially attractive
method to implement the \among\ constraint.
\begin{mytheorem}
GAC on %$\among([X_1,\myldots,X_n],[d_1,\myldots,d_m],N)$
\among\ is equivalent to HC on the decomposition using \roots.
\end{mytheorem}
\myproof %(Sketch)
Suppose the decomposition into
$\roots([X_1,\myldots,X_n],S,\{d_1,\myldots,d_m\})$ and $|S|=N$ is
HC. The variables $X_i$ divide into three categories: those whose
domain only contains elements from $\{d_1, \myldots, d_m\}$ (at
most $\min(N)$ such variables); those whose domain do not contain
any such elements (at most $n-\max(N)$ such vars); those whose
domain contains both elements from this set and from outside.
Consider any value for a variable $X_i$ in the first such
category. To construct support for this value, we assign the
remaining variables in the first category with values from $\{d_1,
\myldots, d_m\}$. If the total number of assigned values is less
than $\min(N)$, we assign a sufficient number of variables from
the second category with values from $\{d_1,\myldots,d_m\}$ to
bring up the count to $\min(N)$. We then assign all the remaining
unassigned $X_j$ with values outside $\{d_1,\myldots,d_m\}$.
Finally, we assign $\min(N)$ to $N$. Support can be constructed
for variables in the %second and third
other two categories in a similar way, as well as for any value of
$N$ between $\min(N)$ and $\max(N)$. \myqed

\subsubsection{At most and at least}

The \atmost\ and \atleast\ constraints are closely related. The
\atmost\ constraint puts an upper bound on the number of variables
using a particular value, whilst the \atleast\ %constraint
puts a lower bound.
For instance, $\atmost([X_1,\myldots,X_n],d,N)$ holds
iff $|\{ i \ | \ X_i=d\}| \leq N$.
Both \atmost\ and \atleast\ can be decomposed
into \roots\ constraints. For example:
\begin{eqnarray*}
& & \atmost([X_1,\myldots,X_n],d,N)  \ \ \myiff    \\
& & \roots([X_1,\myldots,X_n],S,\{d\}) \ \And \ |S| \leq N
\end{eqnarray*}
Again it is polynomial to enforce HC on these cases of the \roots\
constraint, and the decomposition does not hinder propagation.
Decomposition is therefore also a potential method to implement
the \atmost\ and \atleast\ constraints in case we do not have such
constraints available in our constraint toolkit.
\begin{mytheorem}
GAC on %$\atmost([X_1,\myldots,X_n],d,N)$ is equivalent to
\atmost\ is equivalent to
HC on the decomposition.
$\roots([X_1,\myldots,X_n],S,\{d\})$ and
on $|S|\leq N$.

GAC on %$\atleast([X_1,\myldots,X_n],d,N)$ is equivalent to
\atleast\ is equivalent to
HC on the decomposition.
$\roots([X_1,\myldots,X_n],S,\{d\})$ and
on $|S|\geq N$.
\end{mytheorem}
\myproof
%(Sketch)
The proof of the last theorem can be
easily adapted to these two constraints.
\myqed

\subsection{Applications of \range\ and \roots\ constraints}

Some global constraints need both \range\ and \roots\ constraints in
their specifications.

\subsubsection{Assign and number of values}

In bin packing and knapsack problems, we may wish to assign both a
value and a bin to each item, and place constraints on the values
appearing in each bin. For instance, in the steel mill slab design
problem (prob038 in CSPLib), we assign colours and slabs to orders
so that there are a limited number of colours on each slab.
$\assignnvalues([X_1,\myldots,X_n],[Y_1,\myldots,Y_n],N)$ holds
iff $|\{Y_i \ | \ X_i = j\}| \leq N$ for each $j$
\cite{beldiceanu3}. We  cannot expect to enforce GAC on such a
constraint as it is NP-hard to do so in general.

\begin{mytheorem}
Enforcing GAC on
\assignnvalues\ is NP-hard.
\end{mytheorem}
\myproof
Deciding if the
constraint \atmostnvalue\ has a solution is NP-complete, where
$\atmostnvalue([Y_1,\myldots,Y_n],N)$ holds iff
$|\{ Y_i \ | \ 1 \leq i \leq n\}|\leq N$
\cite{comicsNVALUE,comicsCONSTRAINTS06}.
The problem of the existence of a solution in this constraint is
equivalent to  the problem of the existence of a solution in
$\assignnvalues([X_1,\myldots,X_n],[Y_1,\myldots,Y_n],N)$ where
$D(X_i)=\{0\}, \forall i\in 1..n$. Deciding whether \assignnvalues\ is
thus NP-complete and  enforcing GAC is NP-hard.
\myqed

\assignnvalues\ can be decomposed into a set of \range\ and \roots\
constraints:

%\vspace{-0.7em}
\begin{eqnarray*}
& & \assignnvalues([X_1,\myldots,X_n],[Y_1,\myldots,Y_n],N)  \ \ \myiff    \\
& & \forall j \ . \ \roots([X_1,\myldots,X_n],S_j,\{j\}) \ \And \\
& & \hspace{2em} \range([Y_1,\myldots,Y_n],S_j,T_j) \ \And \ |T_j| \leq N
\end{eqnarray*}
%\vspace{-1em}

However, this decomposition hinders propagation.

\begin{mytheorem}
GAC on \assignnvalues\ is stronger than HC on the decomposition.
\end{mytheorem}
\myproof Consider $N=1$, $X_1, X_2 \in \{0\}$, $Y_1\in \{1,2\}, Y_2 \in
\{2,3\}$. HC on the
decomposition enforces  $S_0=\{1,2\}$ and
$\{\} \subseteq T_0 \subseteq \{1,2,3\}$
but no pruning on the $X_i$ and $Y_j$.
However, enforcing GAC on $\assignnvalues([X_1,X_2],[Y_1,Y_2],N)$
prunes 1 from $Y_1$ and 3 from  $Y_2$.
\myqed

\subsubsection{Common}

A generalization of the \among\ and \alldifferent\ constraints
introduced in \cite{beldiceanu3} is the \common\ constraint.
$\common(N,M,[X_1,\myldots,X_n],[Y_1,\myldots,Y_m])$ ensures
$N=|\{i \ | \ \exists j, X_i=Y_j\}|$ and $M=|\{j \ | \ \exists i,
X_i=Y_j\}|$. That is, $N$ variables in $X_i$ take values in common
with $Y_j$ and $M$ variables in $Y_j$ takes values in common with
$X_i$. We prove that we cannot expect to enforce GAC on such a
constraint as it is NP-hard to do so in general.
\begin{mytheorem}
Enforcing GAC on %$\disjoint([X_1,\myldots,X_n],[Y_1,\myldots,Y_m])$ is
\common\ is NP-hard.
\end{mytheorem}
\myproof We again use a transformation from 3-SAT.
Consider a formula $\varphi$ with $n$ Boolean variables and $m$
clauses. For each Boolean variable $i$, we introduce a variable
$X_i \in \{ i, -i\}$. For each clause $c_j=x \vee \neg y \vee z$,
we introduce $Y_j \in \{ x,-y,z\}$. We let $N \in
\{0,\myldots,n\}$ and $M=m$. $\varphi$ has a model iff the
\common\ constraint has a solution
%satisfying assignment
in which the $X_i$ take the literals true in this model. \myqed

One way to propagate a \common\ constraint is to decompose it into
\range\ and \roots\ constraints:

%\vspace{-0.7em}
\begin{eqnarray*}
& & \common(N,M,[X_1,\myldots,X_n],[Y_1,\myldots,Y_m])  \ \ \myiff    \\
& & \range([Y_1,\myldots,Y_m],\{1,\myldots,m\},T) \ \And \\
& & \roots([X_1,\myldots,X_n],S,T) \ \And \ |S|=N \ \And \\
& & \range([X_1,\myldots,X_n],\{1,\myldots,n\},V) \ \And \\
& & \roots([Y_1,\myldots,Y_m],U,V)  \ \And \ |U|=M
\end{eqnarray*}
%\vspace{-1em}

Enforcing HC on this decomposition is polynomial. Decomposition
thus offers a simple method to propagate a \common\
constraint. Not surprisingly, the
decomposition hinders propagation.
\begin{mytheorem}
GAC on \common\ is stronger than HC on the decomposition.
\end{mytheorem}
\myproof Consider $N=M=0$, $X_1, Y_1 \in \{1,2\}$, $X_2, Y_2 \in
\{1,3\}$, $Y_3 \in \{2,3\}$. Hybrid consistency on the
decomposition enforces $\{\} \subseteq T, V \subseteq \{1,2,3\}$,
and $S=U=\{\}$ but no pruning on the $X_i$ and $Y_j$.
%Then
%the decomposition is HC.
However, enforcing GAC on $\common(N,M,[X_1,X_2],[Y_1,Y_2,Y_3])$
prunes 2 from $X_1$, 3 from $X_2$ and 1 from both $Y_1$ and $Y_2$.
\myqed

\subsubsection{Symmetric all different}

In certain domains, we may need to find symmetric solutions. For
example, in sports scheduling problems, if one team is assigned to
play another then the second team should also be assigned to play
the first. $\symalldiff([X_1,\myldots,X_n])$ ensures $X_i = j$ iff
$X_j = i$ \cite{regin99}. It can be decomposed into a set of \range\
and \roots\ constraints:

%\vspace{-0.7em}
\begin{eqnarray*}
& & \symalldiff([X_1,\myldots,X_n])  \ \ \myiff    \\
& & \range([X_1,\myldots,X_n],\{1,\myldots,n\},\{1,\myldots,n\}) \ \land \\
& & \forall i %\in \{1,\myldots,n\}
\, . \, \roots([X_1,\myldots,X_n],S_i,\{i\}) \ \land \ X_i \in S_i
\ \land \ |S_i| =1
\end{eqnarray*}
%\vspace{-1em}

%As before,
It is polynomial to enforce HC on these cases of the \roots\
constraint. However, as with the \alldifferent\ constraint, it is
more effective to use a specialised propagation algorithm like
that in \cite{regin99}.

\begin{mytheorem}
GAC on \symalldiff\ is stronger than HC on the decomposition.
\end{mytheorem}
\myproof Consider $X_1 \in \{2,3\}$, $X_2 \in \{1,3\}$, $X_3 \in
\{1,2\}$, $\{\} \subseteq S_1 \subseteq \{2,3\}$, $\{\} \subseteq
S_2 \subseteq \{1,3\}$, and $\{\} \subseteq S_3 \subseteq
\{1,2\}$. Then the decomposition is HC. However, enforcing GAC on
$\symalldiff([X_1,X_2,X_3])$ will detect unsatisfiability. \myqed

To our knowledge, this constraint has not been integrated into any
constraint solver. Thus, this decomposition provides a means of
propagation for the \symalldiff\ constraint.

%\begin{comment}

\subsubsection{Uses}

In Section \ref{uses}, we decomposed the constraint \uses\ with
\range\ constraints. Another way to propagate a \uses\ constraint
is to decompose it using both \range\ and \roots\  constraints:

%\vspace{-0.7em}
\begin{eqnarray*}
& & \uses([X_1,\myldots,X_n],[Y_1,\myldots,Y_m])  \ \ \myiff    \\
& & \range([X_1,\myldots,X_n],\{1,\myldots,n\},T) \ \And \\
& & \roots([Y_1,\myldots,Y_m],\{1,\myldots,m\},T)
%\ \And \\
%& & T \subseteq S
\end{eqnarray*}
%\vspace{-1em}

Enforcing HC on this decomposition is polynomial. Again, such a
decomposition hinders propagation as achieving GAC on a \uses\
constraint is NP-Hard.
% (the same example in Section~\ref{uses} can
%be used to show this).
%
Interestingly, the decomposition of \uses\ using \range\
constraints presented in Section~\ref{uses} and the decomposition
presented here are equivalent.

\begin{mytheorem}
HC on the decomposition of \uses\ using only \range\ constraints
is equivalent to HC on the decomposition using \range\ and \roots\
constraints.
\end{mytheorem}
\myproof We just need to show that HC on
$\roots([Y_1,\myldots,Y_m],\{1,\myldots,m\},T)$ is equivalent to
HC on $\range([Y_1,\myldots,Y_m],\{1,\myldots,m\},T') \And T'
\subseteq T$. Since, the \range\ and the \roots\ constraints are
over the same set of variables ($[Y_1,\myldots,Y_m]$) and the same
set of indices ($\{1,\myldots,m\}$) is fixed for both, then it
follows that set variable $T'$ maintained by \range\ is a subset
of $T$ maintained by \roots.
\myqed
%\end{comment}

\subsection{Open constraints}

Open global constraints have recently been introduced. They are a new
kind of global constraints   for which the set of variables involved
is not fixed. \range\ and \roots\ constraints 
are particularly useful to specify
many such open global constraints. 

The  \gcc\ constraint has been extended to \opengcc, a
\gcc\ constraint for which the set of variables involved 
is not known in advance \cite{vanHregCPAIOR06}.  
Given  variables $X_1,\myldots,X_n$ and a set variable $S$, 
$\emptyset\subseteq S\subseteq \{1..n\}$, 
$\opengcc([X_1,\myldots,X_n],S,[d_1,\myldots,d_m],[O_1,\myldots,O_m])$
holds iff $|\{ i\in S \ | \ X_i = d_j \}|=O_j$ for all $j$. 
\opengcc\   can be decomposed into
a set of \roots\ constraints in almost the same way as \gcc\ was
decomposed in Section \ref{sec:gcc}: 

\begin{eqnarray*}
& & \opengcc([X_1,\myldots,X_n],S,[d_1,\myldots,d_m],[O_1,\myldots,O_m])  \ \ \myiff
  \\
& & S=\bigcup_{i\in 1..m}S_i \ \And \\
& & \forall i \ . \ \roots([X_1,\myldots,X_n],S_i,\{d_i\}) \ \And
\ |S_i| = O_i
\end{eqnarray*}

Propagators for such an open constraint have not yet been included in 
constraint solvers. 
In \cite{vanHregCPAIOR06}, a propagator is proposed for the case where
$O_i$'s are ground intervals. 
In the decomposition above, the $O_i$'s can either be variables or
ground intervals. 
However, even when $O_i$'s are ground intervals, both  the
decomposition and the propagator presented in \cite{vanHregCPAIOR06}
hinder propagation and are  incomparable to each other.

\begin{mytheorem}
Even if  $O_i$'s are ground intervals,  
%\begin{description}
%\itemsep-0,5ex
%\item 
%(i) HC on the  \opengcc\ constraint is stronger than  the propagator in
%\cite{vanHregCPAIOR06}, 
%\item
(1) HC on the  \opengcc\ constraint is stronger than  HC on
the decomposition using \roots\ constraints, 
%\item
(2) the propagator in \cite{vanHregCPAIOR06} and   HC on
the decomposition using \roots\ constraints are incomparable. 
%\end{description}
\end{mytheorem}
\myproof 
(1) Consider  $X_1, X_2 \in \{1,2\}$, $X_3
\in \{1,2,3\}$, $d_i=i$, $S= \{1,2,3\}$ 
and $O_1, O_2, O_3= [0,1]$.
The decomposition is HC (with $\{\} \subseteq S_1, S_2 \subseteq
\{1,2,3\}$ and $\{\} \subseteq S_3 \subseteq \{3\}$). However,
enforcing HC on the  \opengcc\ constraint will prune 1 and 2 from
$X_3$.  

(2) Consider the example in case (1). The propagator in
\cite{vanHregCPAIOR06} will prune 1 and 2 from
$X_3$ whereas the decomposition is HC. 
Consider  $X_1 \in \{1,2\}$, $X_2\in \{2,3\}$, $X_3
\in \{3,4\}$, $d_i=i$, $\{\}\subseteq S\subseteq \{1,2,3\}$ 
and $O_1=[1,1]$, $O_2=[0,1]$,  $O_3= [0,0]$, $O_4=[0,0]$.
The propagator in \cite{vanHregCPAIOR06} will prune the only value in
the $X_i$ variables which is not HC, that is, value 2 for $X_1$. It
will not prune the bounds on $S$.  
However, enforcing HC on the decomposition using  \roots\ constraints
will set  $S_1=\{1\}$, then will prune value 2 for $X_1$, will shrink
$S_2$ to $\{\} \subseteq S_2\subseteq \{2\}$, will set $S_3=S_4=\{\}$
and will finally shrink $S$ to  $\{1\}\subseteq S\subseteq \{1,2\}$. 
\myqed

As observed in \cite{vanHregCPAIOR06}, the definition of
\opengcc\ subsumes the definition for the open version of the
\alldifferent\ constraint. 
Given  variables $X_1,\myldots,X_n$ and a set variable $S$, 
$\emptyset\subseteq S\subseteq \{1..n\}$, 
$\openalldiff([X_1,\myldots,X_n],S)$
holds iff $X_i\neq X_j,\forall i,j\in S$. 
Interestingly, this constraint can be decomposed using \range\ 
 in almost the same way as \alldiff\ was
decomposed in Section \ref{app:alldiff}. 
 
\begin{eqnarray*}
& & \openalldiff([X_1,\myldots,X_n],S) \ \ \myiff \\
& & \range([X_1,\myldots,X_n],S,T) \ \And\ |S|=|T|
\end{eqnarray*}

Not surprisingly, this decomposition hinders propagation  (see the
example used in Theorem \ref{theorem:alldiff} to show that the decomposition of
\alldiff\ using \range\ hinders propagation). Nevertheless, 
as in the case of \opengcc, we do not know of any polynomial algorithm
for achieving HC on \openalldiff.

\subsection{Applications beyond counting and occurrence constraints}
% [ELEMENT, CHANNELLING, CONTGUITY, ETC]

The \range\ and \roots\ constraints %have been proposed to serve as
%a powerful decomposition tool for
are useful for specifying a wide range of counting and occurrence
constraints. Nevertheless, their expressive power permits their
use to specify many other   constraints. % including \element,

\subsubsection{Element} The \element\ constraint introduced in
\cite{hentenryck88} indexes into an array with a variable. More
precisely, $\element(I,[X_1,\myldots,X_n],J)$ holds iff $X_I=J$.
For example, we can use such a constraint to look up the price of a component
included in a configuration problem. The \element\ constraint can
be decomposed into a \range\ constraint without hindering
propagation:

%\vspace{-0.7em}
\begin{eqnarray*}
& & \element(I,[X_1,\myldots,X_n],J)  \ \ \myiff   \ \ |S|=|T|=1 \ \And \\
& & I\in S \ \And \ J \in T \ \And \
\range([X_1,\myldots,X_n],S,T)
\end{eqnarray*}
%\vspace{-1em}

% [VANH CLAIMS THAT IN HIS ICLP91 PAPER HE HAS A DECOMP THAT DOES NOT
% HINDER PROPAG?]
 \begin{mytheorem}
 GAC on %$\element(I,[X_1,\myldots,X_n],J)$ is equivalent to
 \element\ is equivalent to
 HC on the decomposition.
 %$\range([X_1,\myldots,X_n],S,T)$,
 %$I\in S$, $J \in T$, and BC on
 %$|S|=|T|=1$.
 \end{mytheorem}
 \myproof %(Sketch)
 $S$ has all the values in the domain of $I$ in
 its upper bound.
 Similarly $T$ has all the values in the
 domain of $J$ in its upper bound.
 In addition, $S$ and $T$ are forced to
 take a single value. Thus enforcing HC on
 $\range([X_1,\myldots,X_n],S,T)$ has the
 same effect as enforcing GAC on
 $\element(I,[X_1,\myldots,X_n],J)$.
 \myqed

 \subsubsection{Global contiguity}

The \contiguity\ constraint ensures that, in a sequence of 0/1
variables, those taking the value 1 appear contiguously. This is a
discrete form of convexity. The constraint was introduced in
\cite{maher2002} to model a hardware configuration problem. It can
be decomposed into a \roots\ constraint:

%\vspace{-0.7em}
\begin{eqnarray*}
& & \contiguity([X_1,\myldots,X_n])  \ \ \myiff    \\
& & \roots([X_1,\myldots,X_n],S,\{1\}) \ \And \\
& & X=\max(S) \ \And \ Y=\min(S) \ \And \ |S|=X-Y+1
\end{eqnarray*}
%\vspace{-1em}

% [THERE IS A DECOMP IN TERNARY CONSTRAINTS THAT IS INCOMPARABLE TO THIS
% ONE. SHOULD WE MENTION IT? (SLIDING THE CONSTRAINT THAT SAYS THAT IF
% Xi-1 AND Xi+1 ARE SET TO 1, THEN Xi HAS TO BE SET TO 1)]

% [IN ADDITION, BRAHIM HAS A QUADRATIC DECOMP ACHIEVING GAC: Xi=1 AND
% Xj=1 IMPLIES ALL Xk WITH K IN i..j MUST BE 1]

\noindent Again it is polynomial to enforce HC on this case of the
\roots\ constraint. Unfortunately, decomposition hinders
propagation.
Whilst \range\ and \roots\ can specify concepts quite distant from
counting and occurrences like convexity, it
seems that we may need other algorithmic ideas to propagate them
effectively.
 \begin{mytheorem}
 GAC on %$\contiguity([X_1,\myldots,X_n])$ is stronger than
 \contiguity\ is stronger than
 HC on the decomposition.
 \end{mytheorem}
 \myproof Consider $X_1, X_3 \in \{0,1\}$, $X_2, X_4 \in \{1\}$. Hybrid
 consistency on the decomposition will enforce
 $\{2,4\} \subseteq S \subseteq\{1,2,3,4\}$, $X
 \in \{4\}$, $Y \in \{1,2\}$ and
 $|S|$ to be in $\{3,4\}$ but no pruning will happen. However, enforcing GAC on
 $\contiguity([X_1,\myldots,X_n])$ will prune 0 from $X_3$. \myqed

\section{Experimental results}
\label{sec:exp}

We now experimentally assess the value of using the \range\ and \roots\
constraints in specifying global counting and occurrence
constraints.
For these experiments, we  implemented an  algorithm  achieving
HC on \range\  and an  algorithm achieving HC on the decomposition
of \roots\ presented in Section \ref{sec:roots:algo}. 
Note that our algorithm for the decomposition of \roots\ does
\emph{not} use the  Ilog
Solver primitives  
${\tt member}(value, set)$ and ${\tt not\-member}(value,set)$ because
Ilog Solver  does not appear to give  complete propagation on 
combinations of such primitives (see the discussion in Section \ref{sec:imp}). 
We therefore implemented our own algorithms from scratch.

\subsection{Pruning power of \roots}

In Section  \ref{sec:roots:algo} we proposed a decomposition of the
\roots\ constraint into simple implications.
The purpose of this subsection is to measure the pruning
power of HC on the decomposition of \roots\  with respect to HC on the
original \roots\ constraint when we do not meet any of the conditions
that make HC on the decomposition equivalent to HC on the original
constraint (see Section \ref{sec:roots:poly}). We should bear in mind
that  enforcing HC on the \roots\ constraint is NP-hard in general.
In  order to
enforce HC on the \roots\ constraint we used a simple table
constraint (i.e., a constraint in extension)
that has an exponential time and space complexity.
Consequently, the
size of the instances on which we were able to run
this filtering method was severely constrained.

An instance is a set of integer variables $\{X_1,..X_n\}$ and two
set variables $S$ and $T$. It can be described by a tuple $\langle
n, m, k, r \rangle$. The parameter $n$ stands for the number of
integer variables. These $n$ variables  are  initialised with the
domain $\{1,\ldots,m\}$. The upper bound of $S$ is initialised with
$\{1,\ldots,n\}$ and the upper bound of $T$ is initialised with
$\{1,\ldots,m\}$. The parameter $k$  corresponds to the number of
elements of the set variable $S$ (resp. set variable $T$) that are,
with equal probability, either put  in the lower bound or {excluded}
from the upper bound of $S$ (resp. of $T$). Finally, the parameter
$r$ is the total number of values removed, with uniform
probabilities from the domains  of the integer variables, keeping at
least one value per domain. We generated $1000$ random instances for
each combination of $n, m \in [4,..6]$, $k \in [1..min(n,m)]$ and $r
\in [1..n(m-1)]$.

For each one of the instances we generated, we propagated
\roots$([X_1,..X_n],S,T)$ using either the table constraint
(enforcing HC), or our decomposition (enforcing HC in special
cases). We observed that on $29$ out of the $32$ combinations of the
parameters $n$, $m$ and $k$, the decomposition achieves HC for all
$1000$ instances of every value of $r$. On the remaining three
classes ($\langle 4, 6, 3, * \rangle$, $\langle 5, 6, 3, * \rangle$
and $\langle 6, 6, 3, * \rangle$), the decomposition fails to detect
$0.003\%$ of the inconsistent values.

As a second experiment, we used  the same instances expect that we did
not fix or remove $k$ values randomly from $T$,
that is, in all instances,  $lb(T)=\emptyset$ and $ub(T)=\{1,\ldots,m\}$.
All other settings remained equal.
By doing so, we allowed the random
domains to reach situations equivalent to that
of the  counter example given in the proof
of Theorem~\ref{hinderhc}.
With this setting,
we observed that the decomposition still achieves HC on $18$ out of
the $32$ combinations of the parameters $n$, $m$ and $k$,
for all $1000$ instances of every value of $r$.
On the remaining classes,
the percentage of  inconsistent values not pruned by the decomposition
increases to $0.039\%$.

Clearly, this experiment is limited in its scope, first
by the relatively small size of the instances, and second
by the choices made for generating random domains.
However,
we conclude that examples
of inconsistent values not being detected by the decomposition
appear to be
%relatively
rare.

\subsection{Pruning power and efficiency of \range}

%\chr{effectiveness or pruning power?}

Contrary to  the \roots\ constraint, we have a complete
HC propagator for the \range\ constraint. Thus, we do not need to
assess the pruning power of our propagator. Nevertheless, it can be
interesting to compare the pruning power and the efficiency of
decomposing a global  constraint using \range\ or using another
decomposition with simpler constraints.

The purpose of this subsection is to compare the decomposition of
\uses\ using \range\ constraints against a simple decomposition using
more elementary constraints.
We chose   the \uses\ constraint because  it is NP-hard to achieve GAC
on the \uses\ constraint  (see Section \ref{uses})  and there is no
propagator available for this constraint in the literature.
Furthermore, one of the time-tabling problems at the University of
Montpellier can easily be modelled as a CSP with
\uses\ constraints. 
We first  compare the two decompositions of
\uses\ (with or without \range) in terms of run-time
as well as pruning power on random CSPs. 
Then, we solve the problem of building the set of courses in the
Master of Computer Science at the University of Montpellier with the two
decompositions. 

\subsubsection{Random CSPs}\label{sec:expe:uses:random}
In order to isolate the effect of the \range\ constraint from
other modelling issues, we used the following protocol: we
randomly generated instances of binary CSPs and we added
\uses($[X_1,\myldots,X_n],[Y_1,\myldots,Y_n]$) constraints.
In
all our experiments, we encode \uses\ in two different ways:
\newcommand{\rangem}{\texttt{range}}
\newcommand{\decomp}{\texttt{decomp}}
\begin{description}
\item[\texttt{[}\rangem\texttt{]}:] by decomposing \uses\ using \range\ as
  described in Section \ref{uses}% and using the algorithm
,
\item[\texttt{[}\decomp\texttt{]}:] by decomposing the \uses\
constraint   using primitive constraints as described next.
\end{description}

$$
\uses([X_1,\myldots,X_n],[Y_1,\myldots,Y_n]) \ \ \myiff
$$
$$
\hspace*{3cm} i \in S \rightarrow X_i \in T~\wedge~ j \in T
\rightarrow \exists i \in S. X_i=j  \wedge ~~~~~~
$$
$$
\hspace*{3cm} i \in S' \rightarrow Y_i \in T'~\wedge~ j \in T'
\rightarrow \exists i \in S'. Y_i=j  \wedge ~~~~~~
$$
$$
\hspace*{3cm} T \subseteq T'
$$

The problem instances are generated according to model B in
\cite{prosserAIJ96}, and can be described with the following
parameters: the number of $X$ and $Y$ variables $nx$ and $ny$ in
\uses\ constraints, the total number of  variables $nz$, the
domain size $d$, the number of binary constraint $m_1$, the number
of forbidden tuples $t$ per binary constraint, and the number of
\uses\ constraints
$m_2$. %, and
Note that the \uses\ constraints can have overlapping or disjoint
scopes of variables. We distinguish the two cases.
All reported results are averages on  1000 instances.

\begin{figure}[t]
\begin{center}
%  \subfigure[domain reduction]{
 % \includegraphics[width = .6\textwidth]{pruning.prop.5.10.35.20.70.150.100.3.10.eps}
 \input{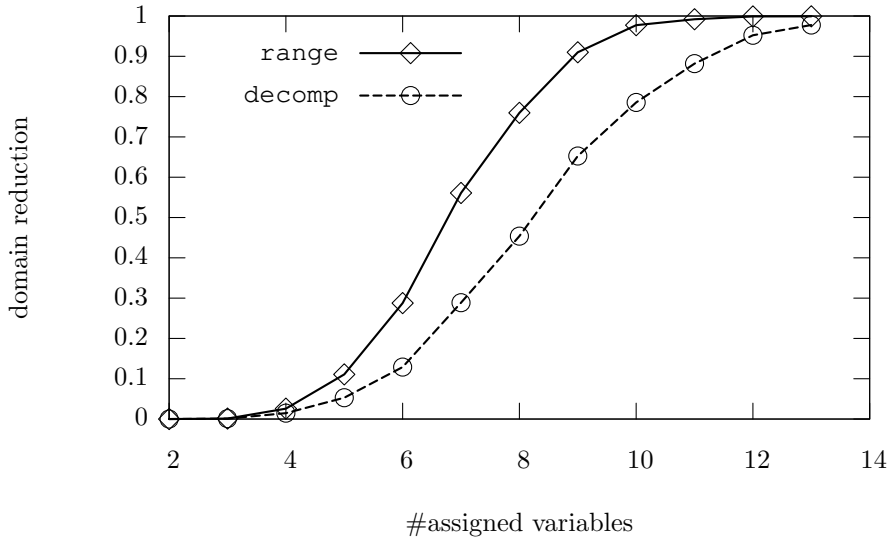}
    %\epsfig{file=avg5.eps, scale=0.4}
%  }
%   \hspace{.2cm}
%   \subfigure[cputime]{
%     \includegraphics[width = .4\textwidth]{time.prop.5.10.35.20.70.150.100.3.10.eps}
%     %\epsfig{file=avg6.eps, scale=0.4}
%   }
\end{center}
\caption{\label{fig:pro:ove} Propagating random binary constraint
satisfaction problems with three overlapping {\sc Uses}
constraints (class A).}
\end{figure}

Our first experiment  studies the effectiveness of decomposing
\uses\ with \range\ for propagation \emph{alone} (not solving). We
compared the number of values removed by propagation on the models
obtained by representing \uses\ constraints in two different
ways, either using \range\ (\rangem) or using the simple decomposition (\decomp).
To simulate what happens inside a backtrack search, we
repeatedly and randomly choose a variable, assign it
to one of its values and propagate the set of random binary constraints.
After doing so for a given number of variables, if the
CSP is still consistent, we enforce HC on  each one of the two
decompositions above.
Hence, in the experiments, the
constraints are exposed to a wide range of different variable
domains. We report the ratio of values removed by propagation on
the following classes of problems:
\begin{eqnarray*}
  class~A: &\langle nx=5, ny=10, nz=35, d=20, m_1=70, t=150, m_2=3\ (overlap) \rangle\\
  class~B: &\langle nx=5, ny=10, nz=45, d=20, m_1=90, t=150, m_2=3\ (disjoint)  \rangle
\end{eqnarray*}
in which the number of assigned variables  varies between 1 and
14.
% Note that the ratio of values removed is given by the
% following formula, where $D_0$ and $D$ respectively denote the
% domain before and after propagation:
% $$
% \sum_{i \in [1..n]}(|D_0(X)|-|D(X)|)/|D_0(X)|
% $$
A failure detected by  the propagation algorithm yields a ratio of 1 (all
values are removed).

\begin{figure}[t]
\begin{center}
%   \subfigure[domain reduction]{
  \input{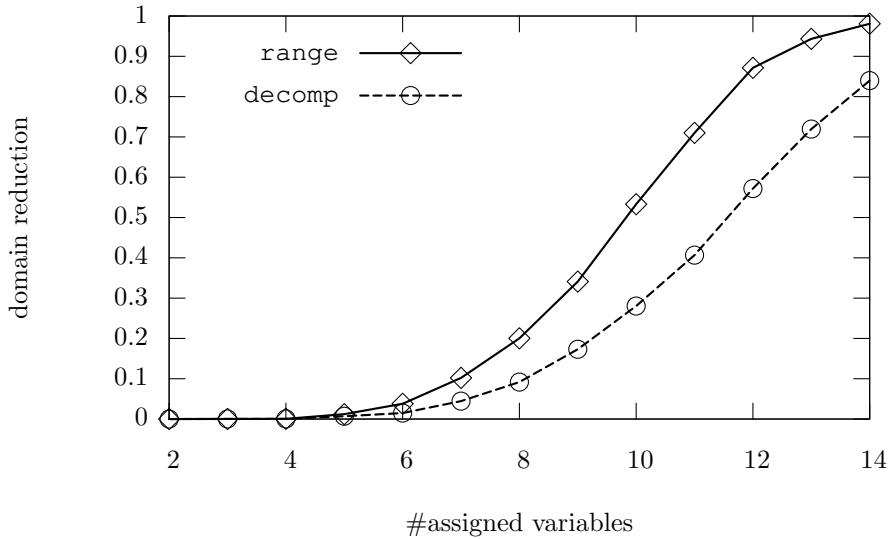}
  %\includegraphics[width = .6\textwidth]{pruning.prop.5.10.45.20.90.150.100.3.15.eps}
    %\epsfig{file=avg5.eps, scale=0.4}
%   }
%   \hspace{.2cm}
%   \subfigure[cputime]{
%     \includegraphics[width = .4\textwidth]{time.prop.5.10.45.20.90.150.100.3.15.eps}
%     %\epsfig{file=avg6.eps, scale=0.4}
%   }
\end{center}
\caption{\label{fig:pro:dis} Propagating random binary constraint
satisfaction problems with three disjoint {\sc Uses} constraints
(class B).}
\end{figure}

We observe in Figures \ref{fig:pro:ove} and \ref{fig:pro:dis} that
 propagating
the \uses\ constraint using the \range\ constraint\ (\rangem\
model) is much more effective than propagating it using the
decomposition using elementary constraints (\decomp\ model).
In certain cases, the \rangem\ model more than doubles the
amount of values pruned. For instance after $7$ random assignments
the \decomp\ model prunes only 28.8\% of the values for the first
problem class (Fig. \ref{fig:pro:ove}) and 4.4\% for the
second (Fig. \ref{fig:pro:dis})
whilst the \range\ algorithm respectively prunes 56\% and 10.2\% of
the values.
%are greater when the \uses\ constraints of the
%original problem overlap (Fig. \ref{fig:pro:ove}) than when they
%are all disjoint (Fig. \ref{fig:pro:dis}).
As we see in the next experiments, such a difference
in pruning can map to considerable
savings when solving a problem.

Our second experiment studies the efficiency of decomposing \uses\
with \range\ when \emph{solving} the problems. Our solver used the
\emph{smallest-domain-first} variable ordering heuristic with the
lexicographical value ordering and a cut-off at 600 seconds. We
compared the cost of solving  the two types of models: %\usesm,
\rangem\ and \decomp. We report the number of fails and the
cpu-time needed to find the first solution on the following
classes of problems:
\begin{eqnarray*}
  class~C: &\langle nx=5, ny=10, nz=25, d=10, m_1=40, t, m_2=2 \rangle\\
  class~D: &\langle nx=5, ny=10, nz=30, d=10, m_1=60, t, m_2=2 \rangle
\end{eqnarray*}
in which $t$ varies between 30 and 80.

\begin{figure}[t]
\begin{center}
%  \subfigure[fails]{
    \includegraphics[width = .6\textwidth]{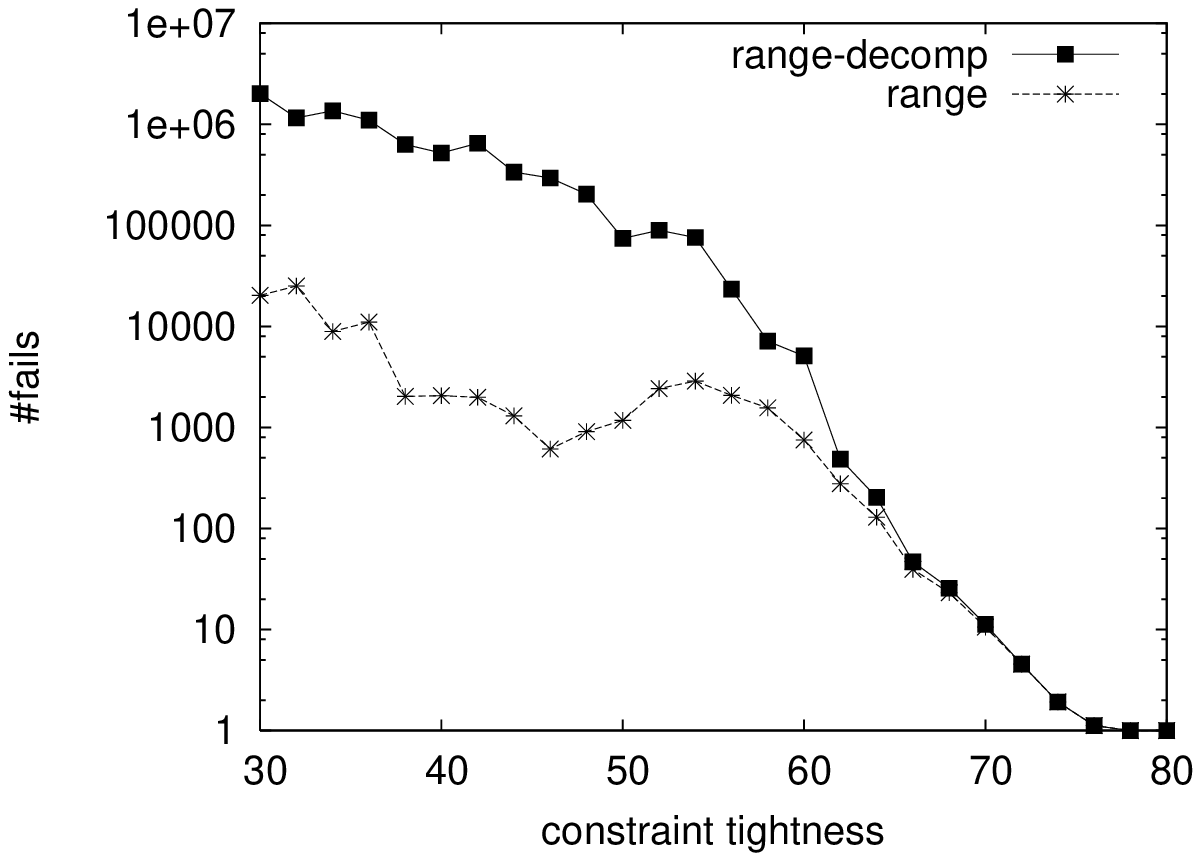}
    %\epsfig{file=avg5.eps, scale=0.4}
%  }
%  \hspace{.2cm}
%  \subfigure[cputime]{
    \includegraphics[width = .6\textwidth]{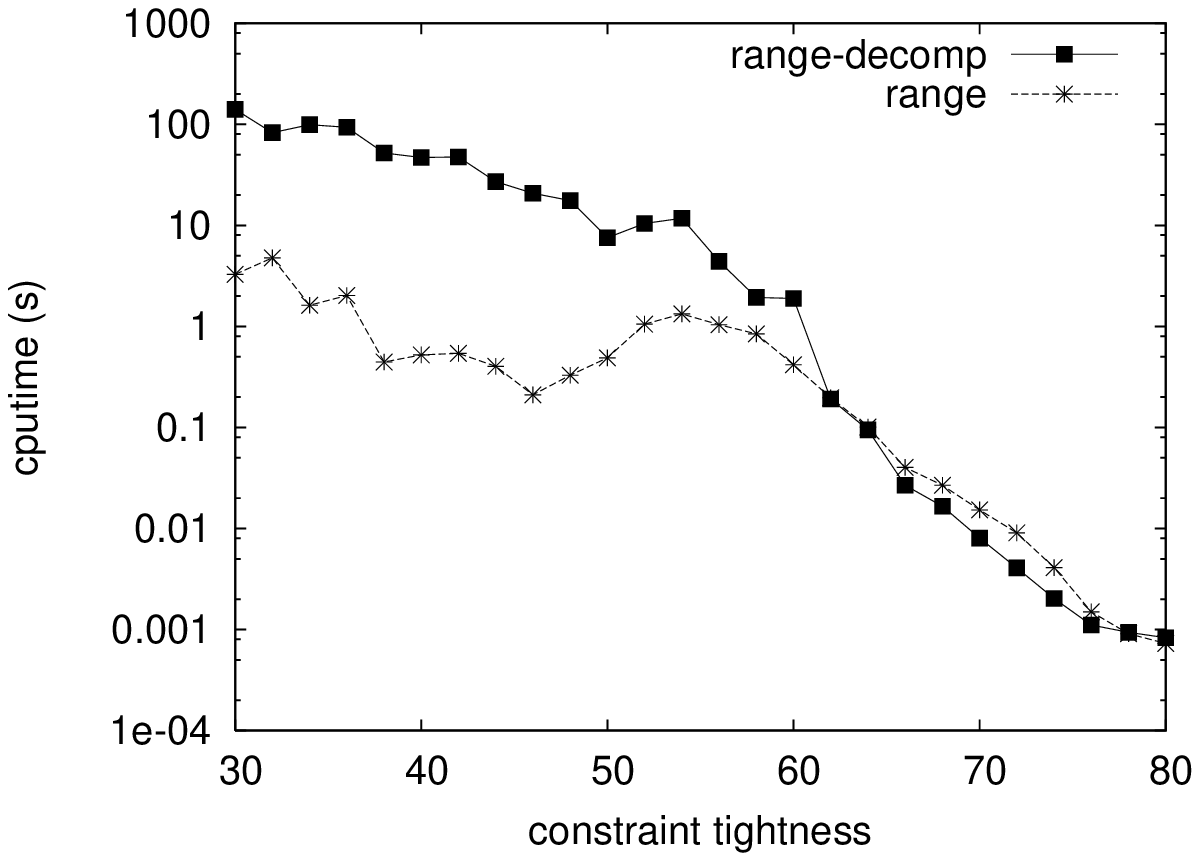}
    %\epsfig{file=avg6.eps, scale=0.4}
%  }
\end{center}
\caption{\label{fig:sol:ove} Solving random binary constraint
satisfaction problems with two overlapping {\sc Uses} constraints
(class C).}
\end{figure}

\begin{figure}[t]
\begin{center}
%  \subfigure[fails]{
    \includegraphics[width = .6\textwidth]{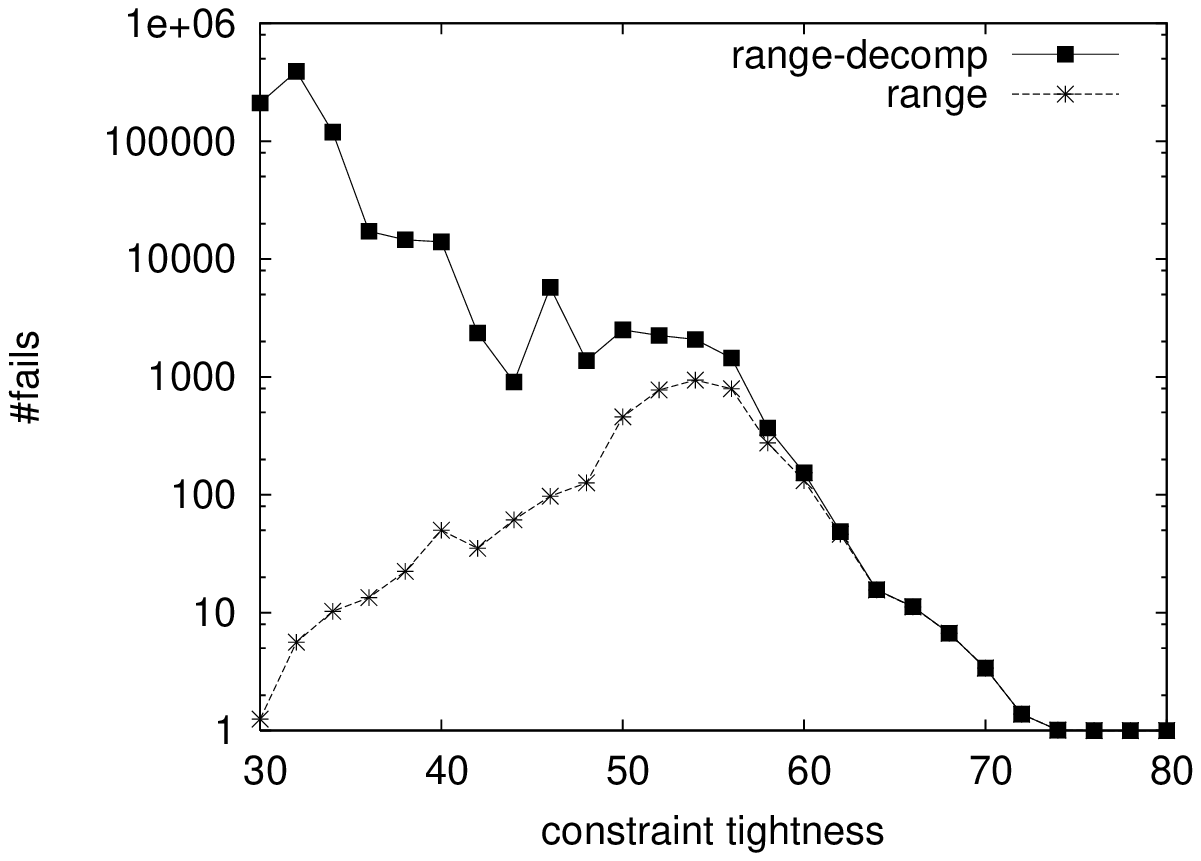}
    %\epsfig{file=avg5.eps, scale=0.4}
%  }
%  \hspace{.2cm}
%  \subfigure[cputime]{
    \includegraphics[width = .6\textwidth]{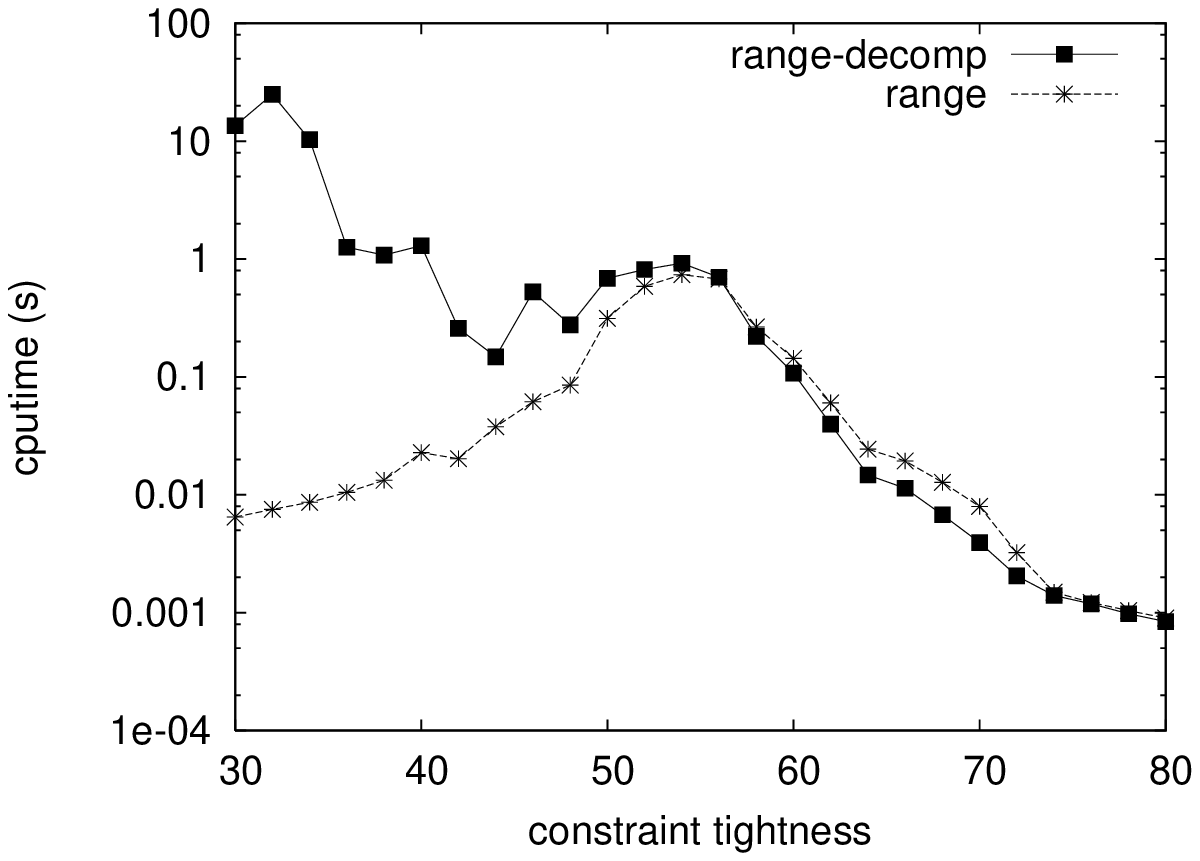}
    %\epsfig{file=avg6.eps, scale=0.4}
%  }
\end{center}
\caption{\label{fig:sol:dis} Solving random binary constraint
satisfaction problems with two disjoint {\sc Uses} constraints
(class D).}
\end{figure}

We observe in Figures \ref{fig:sol:ove} and \ref{fig:sol:dis} that
%solving large  instances
using the decomposition using the elementary constraints (\decomp\
model) is not efficient (note the log scale).
%This is due to the disjunction in the
%implementation of $\exists$.
The instances solved here
(classes C and D) are much smaller than those used for propagation
(classes A and B). Solving larger instances was impractical.
This second experiment shows %how efficiently
%and effectively
that \range\ can reasonably solve problems containing \uses\
constraints. It also shows the clear benefit of using our
algorithm in preference to the decomposition using elementary constraints %of \range\
over the under-constrained region. As the problems get
over-constrained, the binary constraints dominate the pruning, and
the algorithm has a slight overhead in run-time, pruning the same as
the decomposition using elementary constraints.

\subsubsection{Problem of the courses in the master of computer
  science}\label{sec:master} 

To confirm the results obtained on several types of random instances,
we tackle the problem of deciding which courses  to run in the
Master of Computer Science at the University of Montpellier. 
This problem, which is usually  solved  by hand with the help of an Excel
program, can be specified as follows. 
The second year of the Master of Computer Science advertises a set 
$C$ of possible courses. There is a set $L$ of $n$  lecturers  who have
skills to teach some subset of the courses (between 1 and 9 per lecturer). 
There is a set $S$ of  $m$ students who bid for which courses they
would like to attend (between 6 and 10 bids per student). 
A course runs only if at least  5 students bid for it. 
%Teaching to that master  being considered a favour given to the lecturer,
%As teaching is considered a benefit to the lecturer, 
Every lecturer participates in just one course, but
several lecturers can be assigned to the same course. 
There is also a set $P\subseteq L$ of professors who are in charge
of the course in which they participate.  
The goal is to run enough courses so that all lecturers are
assigned to one course and
all students can attend at  least \emph{one} of the courses for
which they bid. 

The models we used have variables $L_i$ representing  which course is  
taught by lecturer $i$ and variables $S_j$ representing   
one of  the courses student $j$ wants to attend. 
$D(L_i)$ contains all courses lecturer $i$ can teach except those that
received less than 5 bids. 
$D(S_j)$ contains all courses student $j$ has bid for, except those that
received less than 5 bids. 
We put a  constraint 
$\uses([L_1,\ldots,L_n],[S_1,\ldots,S_m])$ and a constraint
$\alldiff(L_{i_1},\ldots,L_{i_p})$ where
$\{L_{i_1},\ldots,L_{i_p}\}=P$. 
Model \rangem\ decomposes \uses\ with
\range, and model \decomp\ decomposes \uses\ with primitive constraints
as described in Section \ref{sec:expe:uses:random}. 

In the  only  instance we could obtain from the university, year-2008, 
there are 50 lecturers, 26 professors, 53 courses, 
%among which 17 received less than the 5 required  bids from students, 
and 177 students. 
We solved year-2008,   both with  model
\decomp\ and with model \rangem. Both models could find a  solution
%set of courses that guarantees that every lecturer  participates to
%the master, every professor is the head of a course, and every
%student attends one of her favourite courses 
in a few milliseconds. 

We modified the two models so that the  satisfaction of the students
is improved. Instead of
trying to satisfy only \emph{one} of their choices, we try now to
satisfy $k$ choices. The models are modified in the following
way. 
We create $k$ copies of  each variable $S_j$, that is,
$S^1_j,S^2_j,\ldots,S^k_j$, with $D(S^i_j)$ containing the same values
as $D(S_j)$ (see above). 
We post
constraints $S^1_j<S^2_j<\ldots<S^k_j$ that break symmetries and
guarantee that  $S^1_j, S^2_j,\ldots,S^k_j$  all take  different values. 
Then, instead of having a single \uses\ constraint, we have
$k$ \uses\ constraints, one on each set 
$S^i_1,S^i_2,\ldots,S^i_m$ of  variables%, $i\in1..k$
: $\uses([L_1,\ldots,L_n],[S^1_1,\ldots,S^1_m])$, $\ldots$, 
$\uses([L_1,\ldots,L_n],[S^k_1,\ldots,S^k_m])$. 
Model \rangem-$k$ decomposes \uses\ with
\range, and model \decomp-$k$ decomposes \uses\ with primitive constraints
as described in Section \ref{sec:expe:uses:random}. 

We solved instance year-2008 with $k=2,3,4,5$. When $k=2$ or $k=3$,
both models find a solution in a few milliseconds, \decomp-$k$ being
slightly faster than \rangem-$k$. 
\rangem-4 finds a solution in 4 fails and 5.83 sec. whereas \decomp-4 was stopped
after 24 hours without finding any solution. 
\rangem-5 and  \decomp-5 were stopped
after 24 hours without finding any solution or proving that none exists. 
This  experiment  shows that it can be effective to 
solve a real-world problem containing  a global constraint like
\uses\ by specifying it with \range\ instead of
using a decomposition with elementary constraints.

\subsection{Solving  problems using \range\ and \roots}\label{sec:mystery}

In Section \ref{sec:master}, we showed how decomposing  a global
constraint with \range\ can be useful to solve a real-world problem. 
In this subsection we study another real-world problem that involves a
greater variety of global constraints, some allowing decompositions with
\range, some others with \roots. More importantly, we will compare 
monolithic propagators  of existing well-known global  constraints with their
decompositions using \range\ and \roots. 
The purpose of this subsection is to see if   solving real-world
constraint problems using \range\ and \roots\ leads to acceptable
performance compared to specialised global constraints and
their propagators.

We used a model for the Mystery Shopper problem \cite{cheng99} due
to Helmut Simonis that appears in CSPLib (prob004). We used the
same problem instances as in \cite{comicsIJCAI05} but perform a
more thorough and extensive analysis. We partition the constraints
of this problem into three groups:

\begin{description} \itemsep=-1pt

\item[Temporal and geographical:] All visits for any week are made by different shoppers.
Similarly, a particular area cannot be visited more than once by
the same shopper.
\item[Shopper:] Each shopper makes
exactly the required number of visits.
\item[Saleslady:]
A saleslady must be visited by some shoppers from at least 2
different groups (the shoppers are partitioned into groups).
\end{description}

The first group of constraints can be modelled by using
\alldiff\ constraints \cite{regin1}, the second can be modelled by
\gcc\ \cite{regin2} and the third by \among\ constraints
\cite{beldiceanu5}.
We experimented with several models using Ilog Solver where these
constraints are either implemented as their Ilog Solver primitives
(respectively, {\tt IloAllDiff}, {\tt IloDistribute}, and a
decomposition using {\tt IloSum} on Boolean
variables) or %\range\ and
as their decompositions with \range\ and \roots.
The decomposition of $\among([X_1,\myldots,X_n],[d_1,\myldots,d_m],N)$
we use is the one presented in \cite{comicsERCIMLNAI06}, that is,
$(B_i=1\leftrightarrow X_i\in[d_1,\myldots,d_m]),\forall i\in 1..n \land
\sum_i B_i=N$.
Note that this decomposition of the \among\ constraint maintains GAC in
theory \cite{comicsERCIMLNAI06}.
This decomposition  can be implemented in many
solvers  using disjunctions of membership constraints:
${\tt or}({\tt notmember}(X_i,[d_1,\myldots,d_m]), B_i= 1)$
and
${\tt or}({\tt member}(X_i, [d_1,\myldots,d_m]), B_i= 0)$.
Unfortunately,  Ilog Solver does not appear to achieve GAC on such disjunctions
of primitives because  the negated membership constraint
${\tt notmember}(X_i, [d_1,\myldots,d_m])$ is activated
only if $X_i$ is instantiated with a value in $[d_1,\myldots,d_m]$
whereas it should be as soon as $D(X_i)\subseteq
[d_1,\myldots,d_m]$.
%This explains the discrepancy between the results.

We report results for the following representative %half of the $2^3$ possible
models:
\begin{itemize}

\item {\tt Alld}-{\tt Gcc}-{\tt Sum} uses only Ilog Solver primitives;

\item {\tt Alld}-{\tt Gcc}-{\sc Roots} where \among\ is encoded using \roots;

\item {\tt Alld}-{\sc Roots}-{\tt Sum} where \gcc\ is encoded using \roots;

\item {\sc Range}-{\tt Gcc}-{\tt Sum} where \alldiff\ is encoded
using \range;

\item {\tt Alld}-{\sc Roots}-{\sc Roots} where \among\ and \gcc\ are encoded using \roots;

\end{itemize}

Note that  \among\  encoded as \roots\ uses the decomposition
presented in Section \ref{sec:among}, the \gcc\ uses the
decomposition presented in Section \ref{sec:gcc}, and \alldiff\
uses the decomposition presented in Section \ref{app:alldiff}.

We study the following important questions:

\begin{itemize}

\item How does the \roots\ decomposition of the \among\ constraint compare to the
\sumc\ decomposition in terms of pruning and run-times?

\item Does the decomposition of \gcc\ using \roots\ lead to a reasonable and acceptable loss in performance?

\item Does the decomposition of \alldiff\ using \range\ lead to a reasonable and acceptable loss in performance?

\item Do we gain in performance by branching on the set variables introduced by
the \roots\ decomposition?

\end{itemize}

To answer the first question, we will compare the model {\tt
Alld}-{\tt Gcc}-{\tt Sum} against the model {\tt Alld}-{\tt
Gcc}-{\sc Roots}. To answer the second question, we will compare
the model {\tt Alld}-{\tt Gcc}-{\tt Sum} against the model {\tt
Alld}-{\sc Roots}-{\tt Sum}. To answer the third question, we will
compare the model {\tt Alld}-{\tt Gcc}-{\tt Sum} against the model
{\sc Range}-{\tt Gcc}-{\tt Sum}. To answer the fourth question, we
will compare {\tt Alld}-{\tt Gcc}-{\tt Sum} against the model {\tt
Alld}-{\sc Roots}-{\sc Roots} that branches on the set variables.

The instances we use in the experiments are generated as follows.
For each number of salesladies $s \in \{10,15,20,25,30,35\}$, we
generate $\lceil(s+2/4)*4\rceil$ shoppers, $4$ visits. Furthermore, to
determine the partitioning of the outlets, we bound the number of
salesladies per outlet between a lower bound and an upper bound
and generate all possible partitions within these bounds. The
number of instances for each class is as follows; for 10
salesladies we have 10 instances, for 15 salesladies we have 52
instances, for 20 salesladies we have 35 instances, for 25
salesladies we have 20 instances, for 30 salesladies we have 10
instances, and for 35 salesladies we have 56 instances.

We also tested two variable and value ordering heuristics:

\begin{itemize}
\item We branch on the variables with  the minimum domain first
and assign values lexicographically. We refer to this as $dom$;

\item We assign a shopper to each saleslady for the first, then
for the second week and so on. This a static variables and value
ordering heuristic. We refer to this as $lex$.
\end{itemize}

However, since $lex$ was consistently better than $dom$ we only
report the results using $lex$.

All instances solved in the experiments use a time limit of 5
minutes. For each class of instances we report the number of
instances solved (\textbf{\#solved}), the average cpu-time in
seconds over all instances solved by the method (\textbf{by
self}), the average cpu-time in seconds over all instances solved
by both methods (\textbf{by all}), the average number of failures
over all instances solved by the method (\textbf{by self}), the
average number of failures over all instances solved by both
methods (\textbf{by all}).

\subsubsection{Among}

\begin{table}[t!]
\begin{scriptsize}
  \begin{center}
 \caption{\label{tab-among-lex}The sum decomposition of  \among\ in the Mystery Shopper
 problem versus the \roots\ decomposition using $lex$ as a branching strategy.}
\begin{tabular}{l|| l|rr|rr|| l|rr|rr|| }
\hline
  & \multicolumn{5}{c||}{ alld-gcc-sum-lex } & \multicolumn{5}{c||}{ alld-gcc-roots-lex } \\
\hline
Size & \#solved &
\multicolumn{2}{c|}{ time (sec.)} &
\multicolumn{2}{c||}{ \#fails }&
\#solved &
\multicolumn{2}{c|}{ time (sec.)} &
\multicolumn{2}{c||}{ \#fails } \\
\hline  &  &by self &by all &by self &by all
& &by self  &by all &by self &by all \\
\hline
10 & 9 & 0.01 & 0.01 & 0.89 & 0.89 & 9 & 0.01 & 0.01 & 0.89 & 0.89 \\
15 & 29 & 0.07 & 0.07 & 431.55 & 431.55 & 29 & 0.07 & 0.07 & 281.90 & 281.90 \\
20 & 25 & 0.02 & 0.02 & 10.60 & 10.60 & 25 & 0.02 & 0.02 & 9.48 & 9.48 \\
25 & 16 & 0.03 & 0.03 & 7.06 & 7.06 & 16 & 0.04 & 0.04 & 7.00 & 7.00 \\
30 & 6 & 0.05 & 0.05 & 50.00 & 50.00 & 6 & 0.07 & 0.07 & 49.67 & 49.67 \\
35 & 31 & 0.23 & 0.23 & 414.68 & 414.68 & 31 & 0.24 & 0.24 & 269.32 & 269.32 \\
\hline
\end{tabular}
  \end{center}
\end{scriptsize}
\end{table}

When branching on the integer variables using %either the $dom$
%(Table \ref{tab-among-dom}) or
$lex$ (Table \ref{tab-among-lex}) strategy, the {\tt Alld}-{\tt
Gcc}-{\sc Roots} model tends to perform better than the {\tt
Alld}-{\tt Gcc}-{\tt Sum} model in terms of pruning (smaller number of
fails). Note that the  Sum decomposition misses some pruning because
of the Ilog Solver propagators used in this decomposition, as
explained at the beginning of Section \ref{sec:mystery}. This
explains the discrepancy. Both models solve the same number of
instances. The results show that in this case of the \among\
constraint, our \roots\ decomposition is as efficient as the
decomposition using elementary Sum constraints. Minor run-time
differences are probably due to the cheaper propagator of Ilog Solver
which achieves less pruning.

\subsubsection{Gcc}

The \gcc\ constraint is one of the most efficient and effective
global constraints available in most constraint toolkits. The
results comparing the {\tt Alld}-{\tt Gcc}-{\tt Sum} model versus
its equivalent (the {\tt Alld}-{\sc Roots}-{\tt Sum} model) where
instead of  \gcc\ constraints we use our decomposition using
\roots\ are shown in %Table~\ref{tab-gcc-dom} and
Table~\ref{tab-gcc-lex}.
We observe that when branching on the
integer variables %either using $dom$ or $lex$
using $lex$, the loss in terms of pruning due to our decomposition
is very low: the difference in number of fails is  less than 5\% on
the hardest instances. This means that our decomposition should
scale well when size and difficulty of problems increases. The
difference in run-times is larger (up to more than one order of
magnitude). This can be explained in part by  the  propagation
algorithms for \range\ and \roots\ that we have implemented in Ilog
Solver. They are far from being optimised,  as opposed to the highly
specialised  native \gcc\ propagator.
Overall, the loss appears to be acceptable. 
Our results show that, for the \gcc\ constraint, the decomposition into \roots\
leads to adequate performance for prototyping. Nevertheless, providing
more efficient propagators for
%\range\ and
\roots\ is an interesting and open issue.

\begin{table}[t!]
\begin{scriptsize}
  \begin{center}
 \caption{\label{tab-gcc-lex}The \gcc\ constraints in the Mystery Shopper
 problem versus the \roots\ decomposition using $lex$ as a branching strategy.}
\begin{tabular}{l|| l|rr|rr|| l|rr|rr|| }
\hline
  & \multicolumn{5}{c||}{ alld-gcc-sum-lex } & \multicolumn{5}{c||}{ alld-roots-sum-lex } \\
\hline
Size & \#solved &
\multicolumn{2}{c|}{ time (sec.)} &
\multicolumn{2}{c||}{ \#fails }&
\#solved &
\multicolumn{2}{c|}{ time (sec.)} &
\multicolumn{2}{c||}{ \#fails } \\
\hline  &  &by self &by all &by self &by all
& &by self  &by all &by self &by all \\
 \hline
10 & 9 & 0.01 & 0.01 & 0.89 & 0.89 & 9 & 0.01 & 0.01 & 1.78 & 1.78 \\
15 & 29 & 0.07 & 0.07 & 431.55 & 431.55 & 29 & 0.43 & 0.43 & 434.38 & 434.38 \\
20 & 25 & 0.02 & 0.02 & 10.60 & 10.60 & 25 & 0.10 & 0.10 & 10.60 & 10.60 \\
25 & 16 & 0.03 & 0.03 & 7.06 & 7.06 & 16 & 0.23 & 0.23 & 38.31 & 38.31 \\
30 & 6 & 0.05 & 0.05 & 50.00 & 50.00 & 6 & 0.43 & 0.43 & 72.33 & 72.33 \\
35 & 31 & 0.23 & 0.27 & 414.68 & 505.48 & 23 & 3.89 & 3.89 & 521.74 & 521.74 \\
\hline
\end{tabular}
  \end{center}
\end{scriptsize}
\end{table}

\subsubsection{Alldifferent}

The \alldiff\ constraint is again one of the most efficient and
effective global constraints available in most constraint
toolkits. The results comparing the {\tt Alld}-{\tt Gcc}-{\tt Sum}
model versus its equivalent (the {\sc Range}-{\tt Gcc}-{\tt Sum}
model) where instead of  \alldiff\ constraints we use our
decomposition using \range\ are shown in %Table~\ref{tab-all-dom} and
Table~\ref{tab-all-lex}. We observe that when branching on the
integer variables  using $lex$ both methods achieve the same
amount of pruning even if we are not in a case where
\alldiff\ constraints are \permutation\ constraints (see
Section \ref{app:alldiff}). This means that even when our
decomposition using \range\ theoretically hinders propagation, it can
in practice achieve GAC.
Concerning run-time efficiency, we observe that both methods solve the
same number of instances.
This is probably a consequence of the good level of pruning achieved
by the decomposition of \alldiff\ using \range.
But  the
{\tt Alld}-{\tt Gcc}-{\tt Sum} model is usually faster, up to one
order of magnitude in the extreme case.
Again, this can be explained in part by our basic implementation of the
\range\ and \roots\ propagators in Ilog Solver,
as opposed to the highly  specialised native \alldiff\ propagator.

\begin{table}[t!]
\begin{scriptsize}
  \begin{center}
 \caption{\label{tab-all-lex}The \alldiff\ constraints in the Mystery Shopper
 problem versus the \range\ decomposition using $lex$ as a branching strategy.}
\begin{tabular}{l|| l|rr|rr|| l|rr|rr|| }
\hline
  & \multicolumn{5}{c||}{ alld-gcc-sum-lex } & \multicolumn{5}{c||}{ range-gcc-sum-lex } \\
\hline
Size & \#solved &
\multicolumn{2}{c|}{ time (sec.)} &
\multicolumn{2}{c||}{ \#fails }&
\#solved &
\multicolumn{2}{c|}{ time (sec.)} &
\multicolumn{2}{c||}{ \#fails } \\
\hline  &  &by self &by all &by self &by all
& &by self  &by all &by self &by all \\
 \hline
10 & 9 & 0.01 & 0.01 & 0.89 & 0.89 & 9 & 0.02 & 0.02 & 0.89 & 0.89 \\
15 & 29 & 0.07 & 0.07 & 431.55 & 431.55 & 29 & 0.18 & 0.18 & 431.55 & 431.55 \\
20 & 25 & 0.02 & 0.02 & 10.60 & 10.60 & 25 & 0.17 & 0.17 & 10.60 & 10.60 \\
25 & 16 & 0.03 & 0.03 & 7.06 & 7.06 & 16 & 0.32 & 0.32 & 7.06 & 7.06 \\
30 & 6 & 0.05 & 0.05 & 50.00 & 50.00 & 6 & 0.57 & 0.57 & 50.00 & 50.00 \\
35 & 31 & 0.23 & 0.23 & 414.68 & 414.68 & 31 & 1.39 & 1.39 & 414.68 & 414.68 \\
\hline
\end{tabular}
  \end{center}
\end{scriptsize}
\end{table}

\subsubsection{Exploiting the set variables}

\begin{table}[t!]
\begin{scriptsize}
  \begin{center}
 \caption{\label{tab-set} Branching on set variables in the Mystery Shoppers Problem}
\begin{tabular}{l|| l|rr|rr|| l|rr|rr|| }
\hline
  & \multicolumn{5}{c||}{ alld-gcc-sum-lex } & \multicolumn{5}{c||}{ alld-roots-roots-set } \\
\hline
Size & \#solved &
\multicolumn{2}{c|}{ time (sec.)} &
\multicolumn{2}{c||}{ \#fails }&
\#solved &
\multicolumn{2}{c|}{ time (sec.)} &
\multicolumn{2}{c||}{ \#fails } \\
\hline  &  &by self &by all &by self &by all
& &by self  &by all &by self &by all \\
 \hline
10 & 9 & 0.01 & 0.01 & 0.89 & 0.89 & 10 & 0.05 & 0.05 & 98.20 & 91.33 \\
15 & 29 & 0.07 & 0.07 & 431.55 & 431.55 & 52 & 0.12 & 0.05 & 102.83 & 23.34 \\
20 & 25 & 0.02 & 0.02 & 10.60 & 10.60 & 35 & 1.30 & 1.25 & 852.14 & 794.20 \\
25 & 16 & 0.03 & 0.03 & 7.06 & 7.06 & 20 & 5.08 & 5.12 & 2218.00 & 2170.12 \\
30 & 6 & 0.05 & 0.05 & 50.00 & 50.00 & 10 & 15.05 & 3.65 & 4476.40 & 1675.33 \\
35 & 31 & 0.23 & 0.23 & 414.68 & 412.24 & 51 & 33.88 & 35.86 & 6111.67 & 6410.14 \\
\hline
\end{tabular}
  \end{center}
\end{scriptsize}
\end{table}

In the previous subsections, we have seen that decomposing global
constraints with \range\ and \roots\ constraints is a viable
approach.
Such decompositions  generally give very small (if any) loss in terms
of pruning and they give  acceptable  run-time performance.
However, we have seen that our basic decomposition using
\roots\ can be slow compared to highly specialised propagators
such as those used by  Ilog Solver for the \gcc\ constraint.
In this subsection, we show that, even without optimising our code, we
can  improve the run-time performance of our decomposition just by
exploiting its internal structure through the extra variables it
introduces.

The decomposition of  global constraints using \range\ and
\roots\ introduces extra set variables. We here explore the
possibility of branching on the set variables as follows.
We branch on the set variables first, then on the
integer variables with min domain once all set variables are
instantiated. We refer to this as $set$.
We compare the best
model that uses the available constraints in Ilog Solver (model  {\tt
  Alld}-{\tt Gcc}-{\tt Sum})  versus the best model that
branches on the set variables (model  {\tt Alld}-{\sc Roots}-{\sc
  Roots}, in which the \among\ and the \gcc\ constraints are expressed
using the \roots\ constraint). Surprisingly, we solve significantly
more instances when branching on the set variables than the model
{\tt Alld}-{\tt Gcc}-{\tt Sum}. But, again, {\tt Alld}-{\tt
Gcc}-{\tt Sum} is a more efficient model when it manages to solve
the instance.

These results are primarily due to the better branching strategy.
However, such a strategy would not be easily implementable without
\roots\ since the extra set variables are part of it. We observe
here that the extra set variables introduced by the \roots\
decomposition may provide new possibilities for branching strategies
that might be beneficial in practice.

These results show that by simply  changing the branching strategy so
that it exploits  the internal structure of the decompositions,
we obtain a significant increase in performance. This gain
compensates the loss in  cpu-time caused by the preliminary nature of our
implementation. %This  noteworthy feature  makes our
%decompositions even more interesting.

\section{Conclusion}
\label{sec:concl}

We have proposed  two global constraints useful in specifying many
counting and occurrence constraints: the \range\ constraint which
computes the range of values used by a set of variables, and the
\roots\ constraint which computes the variables in a set mapping
onto particular values. 
These two constraints capture the notion of image and domain of a
function, making them easy to understand to the non expert in
constraint programming. 
We have shown that these two constraints can easily specify 
counting and occurrence constraints. For example, the open versions of
some well-known global constraints can be specified with \range\ and
\roots. 
Beyond counting and occurrence
constraints, we have shown that the expressive power of \range\ and
\roots\ allows them to specify  many other constraints. 

We have proposed propagation algorithms for these two constraints. Hence,
any global constraint specified using \range\ and
\roots\ can be propagated. %The specification immediately provides a %polynomial
%propagation algorithm for any constraint that can be specified.
%
In some cases, this gives a propagation algorithm which 
achieves GAC on the original global constraint (e.g. the
\permutation\ and \among\ constraints). In other cases, this
propagation algorithm may not make the original constraint GAC, but
achieving GAC is NP-hard (e.g. the \nvalue\ and \common\
constraints). Decomposition is then one method to obtain a
polynomial algorithm. In the remaining cases, the propagation
algorithm may not make the constraint GAC, although specialised
propagation algorithms can do so in polynomial time
(e.g. the \symalldiff\ %\gcc\ %and \alldifferent\
constraint). Our method can still be attractive in this last case
as it provides a generic means of propagation for counting and
occurrence constraints when specialised algorithms have not yet been
proposed or are not available in the constraint toolkit.

We have presented a comprehensive study of the \range\ constraint.
We proposed an algorithm for enforcing hybrid consistency on %propagating
\range.
We also have presented a comprehensive study of  the \roots\
constraint. We proved that propagating completely the
\roots\ constraint is intractable in general. We therefore
proposed a decomposition to propagate it partially. This
decomposition achieves hybrid consistency on the  \roots\
constraint under some simple conditions often met in practice. In
addition, enforcing bound consistency on the decomposition
achieves bound consistency on the \roots\ constraint
whatever conditions hold.

Our experiments show the benefit we can obtain by incorporating
the \range\ and the \roots\ constraints in a constraint toolkit.
First, 
despite being intractable, the \roots\ constraint can be
  propagated using the  decomposition we presented. Even if this
  decomposition
  hinders propagation in theory, our experiments show that it is
  seldom the case in practice.
Second, in the absence of specialised propagation algorithms,
\range\ and \roots\ appear to be a simple and a reasonable method
for propagating (possibly intractable) global
constraints that is competitive to other decompositions into more
elementary constraints. Our experiments show that sometimes we do
better than these other  decompositions either in terms of pruning or in
solution time or both (like the case of the decomposition of the
\uses\ constraint). In addition, compared to highly specialised
propagation algorithms like those for the \alldiff\ and
\gcc\ constraints  in Ilog Solver, 
the loss in performance when using 
\range\ and \roots\ was not great. Thus, if the 
constraint toolkit lacks a specialised propagation algorithm, 
\range\ and
\roots\ offer a quick, easy, and acceptable way of
propagation.
Finally, we observed that the extra set variables
introduced in \range\ and \roots\ decompositions 
can be exploited in the design of new branching
strategies. These extra set
variables may provide both a modelling and solving advantage to
the user.
We hope that by presenting these results, developers of the many
different constraint toolkits will be encouraged to include the
\range\ and \roots\ constraints into their solvers.

%In our future work, we intend to perform more exhaustive empirical
%studies on constraint models  using \range\ and \roots\
%constraints and demonstrate their full applicability in real-life
%problems. We will also consider other classes of global
%constraints (e.g. sequencing constraints) and  identify the
%primitives needed to specify and propagate these.

\section*{Acknowledgements}
We thank Eric Bourreau for having provided the data for the problem of
the master of computer science of the university of Montpellier. We
also thank our reviewers for their helpful  comments which 
improved this paper.

%
%\bibliographystyle{plain}
%\bibliography{biblio}
%

\end{document}